# GraphDecisionBench: Benchmarking System One Models on Graph Tasks

Xianliang Yang[1], Yapu Zhang[2], Li Zhao[1]
[1]Microsoft [2]Beijing University of Technology

**Abstract**

Large language models (LLMs) are increasingly explored for graph understanding and decision-making, while System One models such as Jev select directly from supplied options. However, the capabilities of System One models on graph-related tasks remain unclear. We introduce GraphDecisionBench, a model-independent benchmark that combines structural task profiles, matched graph-text input contrasts and heuristic-proposal controls to diagnose graph decision performance. We evaluate Jev and related choice-based models alongside language-model baselines, covering fourteen model-interface configurations. Jev's results illustrate the benchmark's central distinctions: accurate adjacency recognition does not guarantee broader structural correctness, joint graph-text input does not consistently improve prediction, and feasible construction does not establish high solution quality. Its task contracts, candidate interfaces and scoring rules support comparison across native selectors and language-model adapters. Code and aggregate results are available at https://github.com/VictorYXL/GraphDecisionBench.

## 1 Introduction

Graphs encode relationships among entities. Large language models (LLMs) are increasingly explored for graph-structure questions, node-label prediction, and graph optimization [3, 6, 26, 29, 33]. Sequential applications such as route construction require repeated choices, motivating evaluation of the complete sequence of decisions.

Emerging System One models, such as Jev, select from supplied options and return probabilities to support repeated decisions [1]. We study the correctness and quality of their graph decisions. Structural judgments, graph-text information use and sequential solution construction remain distinct evaluation targets.

Evaluating a model on existing graph tasks is necessary but does not by itself answer three different questions: whether success transfers across structural operations, whether an input component adds value beyond the accompanying evidence, and whether an assisted solution reflects effective model selection rather than strong external proposals. A pooled accuracy or final objective can obscure these distinctions. GraphDecisionBench therefore organizes its evaluation around task-level profiles, matched input conditions and candidate-pool controls, rather than treating the three task families as a single capability score.

We introduce **GraphDecisionBench**, a benchmark addressing three complementary research questions. As shown in Figure 1, GraphDecisionBench includes three main task families: **RQ1: Graph Cognition.** How accurately do models answer questions about local connections and broader graph properties involving paths, cycles, and connectivity? **RQ2: Graph-Text Decision-Making.** For the same classification targets or retrieval queries, how does performance differ between textual input, relational input, and their combination? **RQ3: Sequential Decision-Making on Graphs.** Can successive model choices produce complete, feasible solutions, and how does their quality compare with classical methods and fixed rules? Together, these questions examine how System One models understand graph structure, use graph and text information, and make sequential decisions for graph optimization.

Our contributions are threefold:

**1. A model-independent diagnostic benchmark.** GraphDecisionBench connects structural task profiles, graph-text input contrasts and sequential construction controls through a common choice interface. The task definitions and evaluator are independent of Jev, allowing the same evaluation contracts to be used for other decision models.

**2. Controlled information comparisons.** Matched graph-text conditions assess performance differences from adding explicit edges while holding text and shared label anchors or candidate evidence fixed. The comparison covers both node classification and candidate-based retrieval.

**3. Candidate and solution-quality diagnostics.** Direct-construction, within-pool random and fixed-rule comparisons assess the value of model selection relative to externally supplied heuristic guidance. Across these evaluations, System One and other decision models are compared with LLMs and classical controls.

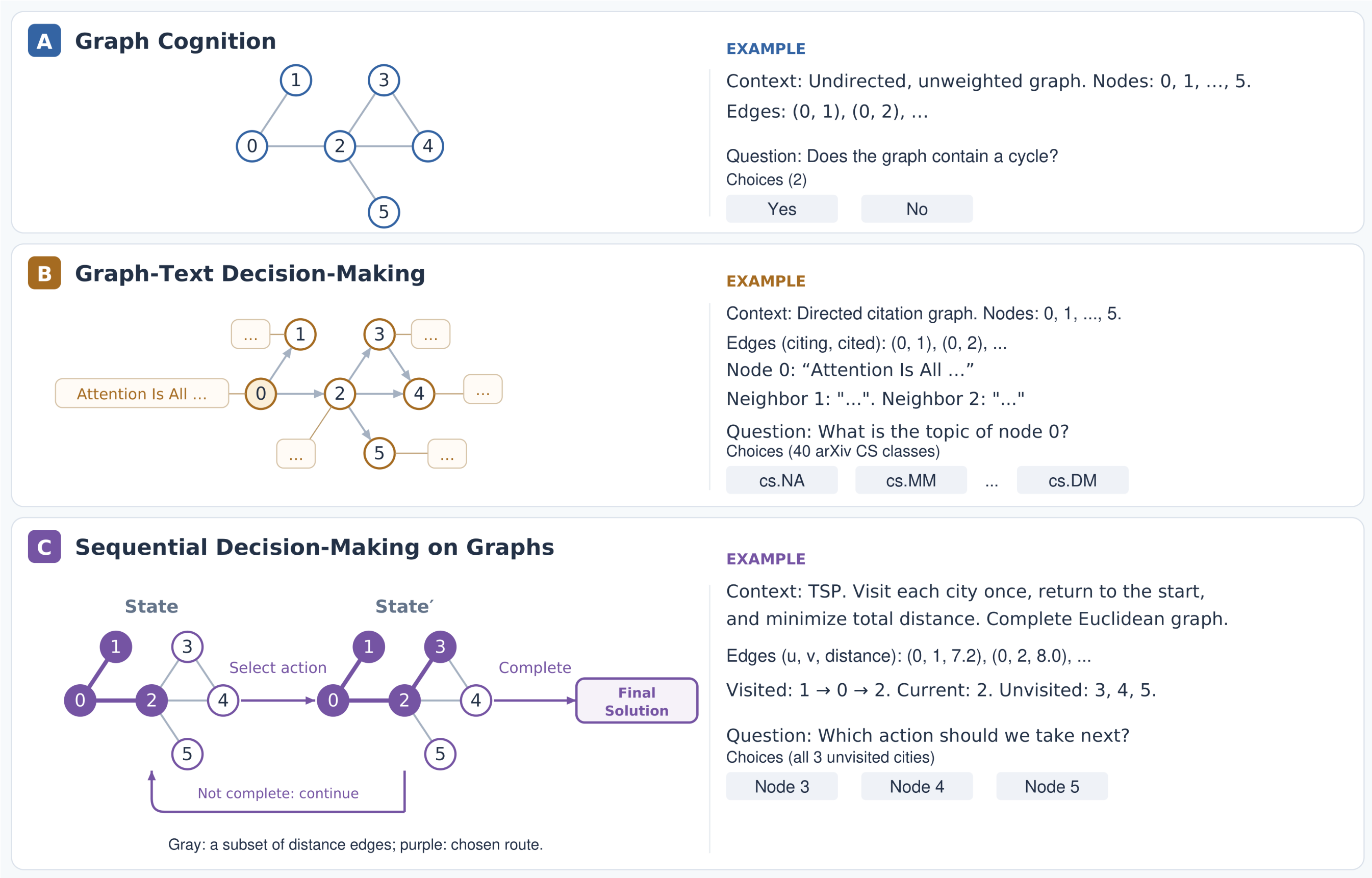

Figure 1. Overview of GraphDecisionBench's three task families. (A) Graph Cognition evaluates structural answer accuracy against algorithmically computed ground truth. (B) Graph-Text Decision-Making evaluates classification and candidate selection under controlled textual and relational inputs; the illustration shows the classification interface. (C) Sequential Decision-Making evaluates TSP, signed MaxCut, deterministic influence maximization, and binary modularity, comparing direct action selection with selection among fixed-heuristic proposals.

# 2 Related Work

## 2.1 Benchmarks for Graph Learning and Reasoning

Graph benchmarks cover prediction, algorithmic reasoning, and computational problem-solving. Open Graph Benchmark provides standardized datasets and evaluation protocols for graph learning [8], while CLRS evaluates the learned execution of algorithms [28]. NLGraph tests graph reasoning through natural-language questions [29], and GraphArena extends evaluation to diverse graph computational problems with explicit assessment of feasibility and solution quality [26]. Beyond structural computation, STaRK evaluates retrieval using textual and relational information in knowledge bases [30].

GraphDecisionBench builds on established graph problems and public data, rather than introducing new optimization objectives. Its emphasis is the combination of diagnostic comparisons: task profiles reveal uneven structural performance, matched input contrasts assess additional evidence, and proposal-pool controls assess selection relative to supplied guidance. These comparisons complement task-level benchmark scores without assuming that the underlying tasks are themselves novel.

## 2.2 Language Models for Graph Understanding and Optimization

GraphWiz [3] and GraphGPT [25] explore graph-specific instruction tuning, while GraphLLM integrates graph representations with language models [2]. For text-attributed graphs, GLEM couples language and graph models [33], while TAPE enriches node representations with LLM-generated explanations [6]. Learned constructive policies [10] and attention-based routing models [11] build solutions through successive decisions. FunSearch [23], EoH [14], ReEvo [32], and HeurAgenix [31] explore LLM-assisted program search, heuristic evolution, and heuristic selection for computational and optimization problems.

### 2.3 System One Decision-Making and Evaluation

Jev targets option selection with associated probabilities [1], while SemIf [13], Decider [15], Kev [18], and Laya [4] provide related decision interfaces. Existing evaluations examine complementary aspects: JevBench studies accuracy, calibration, and cost [24]; WorldTour evaluates Jev-guided route construction [12]; and JevAdvBench examines decision changes under input perturbations relative to repeated-call variability [7]. GraphDecisionBench complements these efforts by evaluating structural correctness, graph-text information utility, and sequential solution quality within a common choice-based framework.

## 3 The GraphDecisionBench Benchmark

GraphDecisionBench separates task specification from model execution. Its graph instances, answer spaces, reference evaluators and scoring rules define the benchmark, while a model adapter supplies decisions. Jev is the main empirical case study, not a component required by the task definitions or evaluator.

We use System One operationally for decision configurations that return a choice without an explicitly requested deliberative reasoning stage. Native choice-based models form the main study group, with no-thinking language-model configurations providing comparison baselines. This evaluation designation does not assert a shared architecture or reveal unobserved internal computation. A choice interface, a no-reasoning setting and low latency are distinct properties; the benchmark assesses decision quality, not a general efficiency advantage.

### 3.1 Tasks and Data Sources

**Graph Cognition (RQ1).** Six query tasks on simple undirected graphs cover adjacency, exact degree, cycle existence, pairwise connectivity, distance thresholds, and articulation points. Induced subgraphs with 8–12 vertices are drawn from Facebook [34], ca-GrQc [35], Power Grid, and Human PPI. The evaluation contains 800 queries: 400 degree queries and eighty for each other task, with 200 queries per source. Queries are stratified by source, size and answer: binary tasks balance yes/no answers, and degree tasks balance the possible degrees within each size. Models answer from the supplied graph, including isolated vertices; independent graph algorithms compute the reference answers. Appendix A.1 defines the predicates and sampling quotas, and Appendix C reports aggregate and source-specific results.

**Graph-Text Decision-Making (RQ2).** A 40-class node-classification task on ogbn-arxiv [8] and a biomedical entity-selection task on STaRK-Prime [30, 36] compare predictions under controlled textual and relational inputs. The primary panels contain 100 test targets and 100 test queries, respectively. ogbn-arxiv conditions share targets, candidate categories and training-label anchors. STaRK-Prime conditions share the query and forty distinct candidate entities, accepting any native correct answer. Table A.5 distinguishes these primary panels from the separate 200-target ogbn-arxiv evaluation.

**Sequential Decision-Making on Graphs (RQ3).** We evaluate four graph-optimization tasks, each using ten public-data instances.

- **TSP.** The model selects unvisited cities sequentially to construct a minimum-length closed tour. We use complete TSPLIB EUC_2D instances with 76–225 cities and their published proven optima as reference values [22].
- **Signed MaxCut.** The model assigns vertices to two sides to maximize the total signed weight of crossing edges. We use OPTSICOM Set2 graphs with 125 vertices, 375 edges and weights $\pm 1$, with historical best-known values from 2016 as references [5, 17].
- **Deterministic LT influence maximization.** The model selects two distinct seeds, activated simultaneously, to maximize the final number of active nodes under fixed-threshold linear-threshold diffusion [9]. We use directed induced subgraphs of SNAP email-Eu-core [37] with 50–150 vertices and exact reference optima obtained by enumerating all seed pairs.
- **Binary modularity maximization.** The model assigns vertices to at most two communities to maximize resolution-1 modularity [16, 38]. We use induced subgraphs of SNAP ca-GrQc [35] with 50–150 vertices and numerical MILP-solver-certified optima as reference values.

Appendix A lists the optimization instances and their data specifications, while Appendix G details the LT and modularity reference calculations and their per-instance values.

### 3.2 Decision Interface and Input Conditions

Each call supplies an input x and candidate set A and returns a choice. A model-interface configuration specifies the model and its readout: native selection, constrained generation or candidate scoring. Compared configurations share task data and prompt-construction rules. For RQ1 and RQ2, the query and condition determine the supplied input and options; RQ3 inputs follow the evolving state described below.

To evaluate another model under the same protocol, an adapter maps the supplied input and candidate set to the model's interface and maps its response back to a candidate identifier. The adapter must preserve candidate identities and the condition-specific evidence, and leave ground-truth answers, objective evaluation and state updates to the evaluator. Native selection, constrained generation and candidate scoring are alternative readouts, not different task definitions. An unsupported request is recorded as unsupported rather than silently reducing the answer space. The adapter and inference settings identify the configuration being compared.

The evaluation unit is a query for RQ1, a target–condition or query–condition pair for RQ2, and a trajectory attempt for RQ3. Each RQ3 trajectory consists of successive decisions made through the task interface.

**Graph-text input conditions.** RQ2 uses five matched conditions, summarized in Table 1. T supplies target or candidate text; G supplies relational evidence; TG combines entity text and relations; BAG is the *edge-ablated joint input*; and A/A* is the comparator without additional entity text or edges. BAG is an input ablation, not Build-a-Graph prompting. In ogbn-arxiv, graph-selected context identities and training-label anchors are shared across conditions, so T is not a graph-independent text baseline.

For STaRK-Prime, every condition retains the query and the same forty candidate identities, names and types. G, TG and BAG also contain the relation-selected context entities. Each candidate set contains at least one native correct answer without marking it in the input: the task evaluates selection within a gold-containing pool, not end-to-end retrieval. TG-BAG removes only explicit edges while preserving the remaining input. G-A* adds relations and their context entities, rather than isolating edges alone. Appendix B gives the candidate-construction details.

Table 1. RQ2 input composition under matched conditions.

| Dataset / condition | Additional entity text | Explicit edges | Other supplied evidence |
|---|---|---|---|
| ogbn-arxiv: T | Target | No | Shared anchors and context IDs |
| ogbn-arxiv: G | None | Directed edges | Shared anchors and context IDs |
| ogbn-arxiv: TG | Target + context | Directed edges | Shared anchors and context IDs |
| ogbn-arxiv: BAG | Target + context | No | Same nodes and text as TG |
| ogbn-arxiv: A | None | No | Shared anchors and context IDs |
| STaRK-Prime: T | Candidates | No | Candidate IDs, names and types |
| STaRK-Prime: G | None | Relations | Candidates + context entities |
| STaRK-Prime: TG | Candidates + context | Relations | Candidates + context entities |
| STaRK-Prime: BAG | Candidates + context | No | Same nodes and text as TG |
| STaRK-Prime: A* | None | No | Candidate IDs, names and types |

All ogbn-arxiv conditions share graph-selected context identities, training-label anchors and forty class options; held-out labels are excluded. All STaRK-Prime conditions share the query and forty candidate entities. Names and types are retained even when additional entity descriptions are absent. Context entities are relation-selected. BAG removes only explicit edges from TG; omitted information is unavailable, not known to be absent.

**Sequential construction modes.** Direct construction supplies all currently legal actions, while Heuristic-proposal selection supplies distinct actions nominated by four named fixed rules. The four tasks, ten instances per task and two modes define eighty trajectory-evaluation units per configuration. Task-specific answer sets and candidate-processing details appear in Appendices A and B.

In RQ3, the state $s_t$ contains the graph, task specification and current partial solution, from which the input $x_t$ and candidate set $A_t$ are constructed. The model selects an action, and the evaluator updates the state:

$$a_t = D_\theta(x_t, \mathcal{A}_t) \in \mathcal{A}_t, \qquad s_{t+1} = F(s_t, a_t). \tag{1}$$

Each selected action advances construction by one step, following the model's own decisions until completion or failure. Different models or construction modes can induce different histories, so later states and candidate sets need not coincide. Automatic completion, candidate rules and exception handling are specified in Appendices A and B.

### 3.3 Evaluation Protocol

Reference answers and optimization reference values are maintained by the evaluator and are not supplied to models as answers. Held-out classification labels are excluded from training-label anchors. Evaluation follows the task-specific protocols below.

**Graph cognition.** The primary score for each task is query accuracy averaged equally across the five graph sizes. We additionally report task-size macro, task macro and query-weighted accuracy, with aggregation definitions in Appendix C. Invalid outputs and failed predictions count as incorrect, while output validity and interface support are reported separately.

**Graph-text decision-making.** We report accuracy and macro-F1 over the fixed forty classes for ogbn-arxiv, and Hit@1 for STaRK-Prime, accepting any native correct answer. We report paired performance differences between the matched input conditions in Section 3.2. Let S(C) denote ogbn-arxiv accuracy or STaRK-Prime Hit@1, expressed as a percentage, under condition C. Define $C_0$ = A for ogbn-arxiv and $C_0$ = A* for STaRK-Prime. The principal contrasts are

$$\Delta_{\text{edge}} = S(\text{TG}) - S(\text{BAG}), \qquad \Delta_{\text{relation}} = S(\text{G}) - S(C_0). \tag{2}$$

Both differences are measured in percentage points. TG-BAG compares explicit edges while holding accompanying text and anchors or candidates fixed. G-$C_0$ compares relational input without additional entity text. We also report TG-T and TG-G; in particular, ogbn-arxiv TG-T adds both context text and edges and is not an edge-only contrast.

Invalid outputs and failed predictions count as incorrect; unsupported interfaces are reported separately. Paired 95% bootstrap intervals preserve the pairing of conditions for each target or query. Statistical details and interpretation are provided in Appendix B.3.

**Sequential decision-making on graphs.** Solution quality is measured by the gap to the reference value for each instance. For a completed instance i, let $f_i$ denote its TSP tour length, signed MaxCut objective or deterministic LT spread, with positive reference $R_i$. Let $Q_i$ and $Q^*_i$ denote the attained and reference binary-modularity values. We define

$$\Delta_i = \begin{cases} \dfrac{|f_i - R_i|}{R_i} \times 100, & \text{TSP, signed MaxCut, or deterministic LT,} \\[2ex] Q_i^{\star} - Q_i, & \text{binary modularity.} \end{cases} \tag{3}$$

TSP, MaxCut and LT gaps are relative deviations reported in percent; multiplication by 100 converts the ratio to a percentage value. Modularity gaps are absolute differences in Q. No completed primary TSP, MaxCut or LT solution improves on its reference, so the absolute-deviation form agrees with the minimization or maximization shortfall on these instances. Lower is better within each task, but gaps across tasks are not directly comparable. $R_i$ is a proven optimum for TSP, an exhaustive optimum for LT, and a historical best-known value for signed MaxCut. $Q^*_i$ is the numerical MILP-certified modularity optimum. A MaxCut gap is therefore a reference gap, not a certified optimality gap. Modularity residuals retain their sign and are not clamped.

For each model configuration, task and construction mode, let N be the number of scheduled trajectory-evaluation units. Define $c_i = 1$ if unit i produces a complete legal solution and $c_i = 0$ otherwise, and let $I = \{i : c_i = 1\}$. We report completion coverage and mean gap over completed trajectories:

$$\text{Coverage} = \frac{1}{N}\sum_{i=1}^{N} c_i, \qquad \mu_{\text{gap}} = \frac{1}{|\mathcal{I}|}\sum_{i \in \mathcal{I}} \Delta_i. \tag{4}$$

Coverage in Eq. (4) is a proportion, displayed as a percentage in result tables. The mean gap is undefined when no trajectory completes. Failed and unsupported units remain in the coverage denominator and are reported separately, without artificial objective values. Paired mode comparisons and comparisons with controls use the same completed-instance subset. Appendix B reports coverage and terminal outcomes. Main-text percentage results are rounded to two decimals; detailed values remain in the appendices.

# 4 Experiments

## 4.1 Experimental Setup

We evaluate Jev [1, 27], Decider-2B v11 [15], Kev-4B [18] and Laya [4] as native choice-based models. No-thinking Qwen3.5-0.8B, 2B, 4B and 9B [20] provide generation baselines, while Qwen3.5-27B, Qwen3.8-27B [21] and Qwen2.5-72B-Instruct [19] extend the comparison to larger models. We also evaluate Qwen3.5-4B through candidate-token scoring to compare decision interfaces. GPT-5.4 without reasoning (reasoning_effort="none") serves as a large-model System One baseline, and GPT-6-Astra, with medium reasoning observed in response metadata, serves as a strong reference. These comparisons comprise fourteen model-interface configurations. Detailed model settings and the GPU candidate-scoring implementation appear in Appendix B.1 and Appendix F.1, respectively. The configuration set covers native choice-based interfaces, generation baselines across model sizes, an alternative scoring readout and a reasoning-enabled reference. This selection targets complementary decision configurations rather than an exhaustive model census; interface limitations remain visible instead of changing the benchmark to accommodate a particular model.

## 4.2 Graph Cognition (RQ1)

We examine whether accurate local connection recognition extends to other structural judgments. Figure 2 presents task-level accuracy, highlighting the capability profiles of Jev and related choice-based models against language-model baselines.

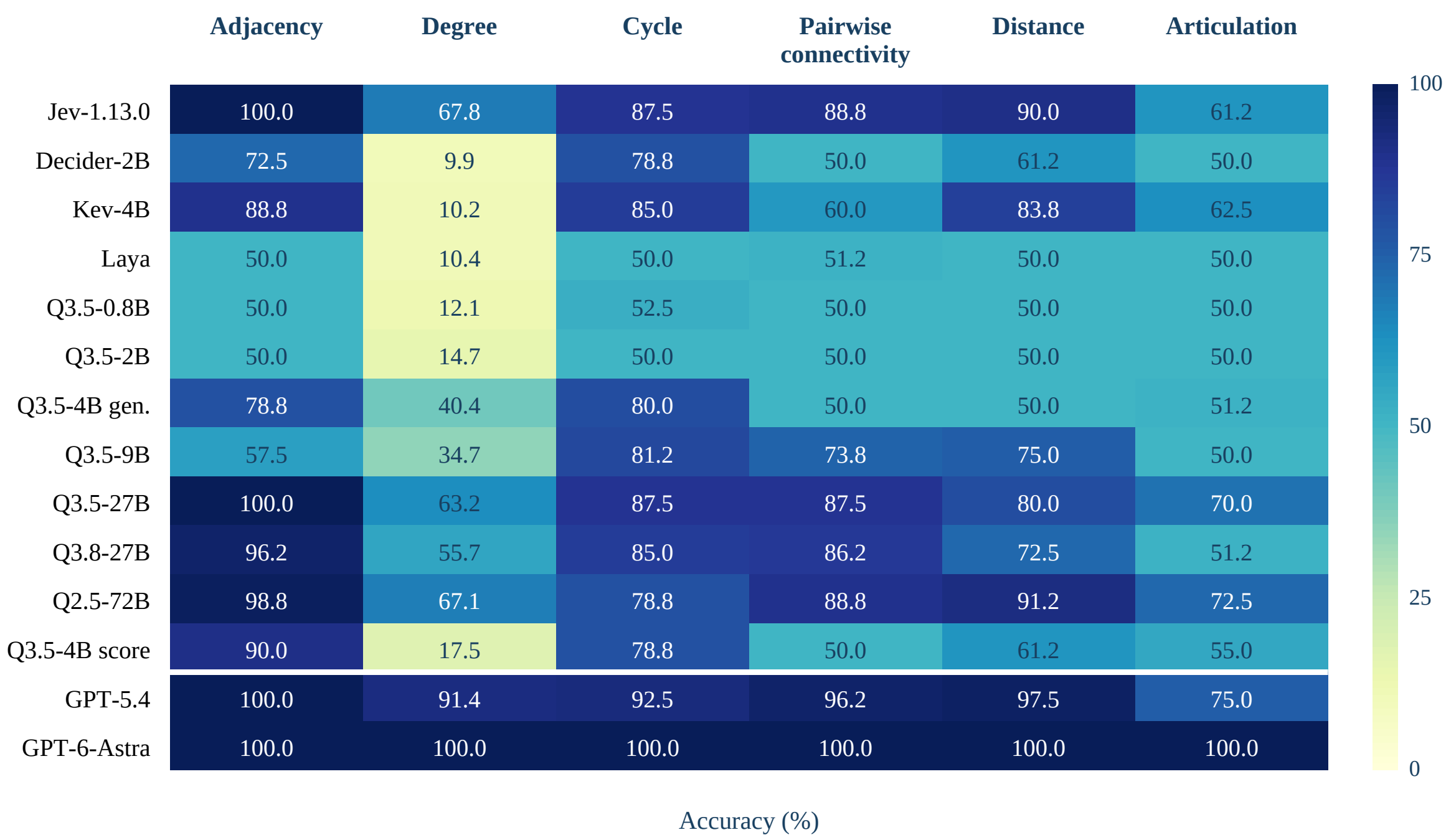

Figure 2. Graph-cognition accuracy across six tasks and fourteen configurations. Values are percentages, equally averaged over sizes 8–12 within each task; higher is better. Pairwise connectivity asks whether a specified vertex pair is connected, not whether the whole graph is connected. Degree is multiclass; all other tasks are binary. Appendix C provides aggregate and source-specific results.

Jev exhibits a strong but uneven structural capability profile. It answers every adjacency query correctly and performs relatively well on cycle, pairwise connectivity and distance judgments, but reaches only 67.80% on exact degree and 61.25% on articulation. Recognizing an explicitly listed connection therefore does not establish accurate neighbor aggregation or identification of a vertex whose removal increases the component count. The contrast locates Jev's limitations at particular structural operations rather than at graph input handling in general; it does not identify the model's internal reasoning mechanism.

Related choice-based models share this task dependence but not Jev's overall level of performance. Decider and Kev answer some binary structural questions relatively well, yet their exact-degree accuracies are only about 10%. The common selection interface thus does not imply comparable structural competence. Jev exceeds both on five of the six task scores, but Kev's articulation accuracy is slightly higher (62.50% versus 61.25%), ruling out a claim of uniform dominance. Laya supports these inputs but records approximately 50% on the binary tasks; these accuracies alone do not establish that it selects randomly.

The language-model baselines place these results in context. Jev outperforms the evaluated smaller Qwen configurations across the six task scores and is competitive with the larger Qwen models, while remaining below the high, broad accuracy of the GPT baselines. Its comparison with larger Qwen models is task-specific: Jev matches Qwen3.5-27B on adjacency and cycle detection, whereas Qwen2.5-72B is stronger on distance and articulation. A small aggregate ordering reversal between Jev and Qwen2.5-72B under query-weighted versus task-size-macro scoring further favors interpreting the task profile rather than a single overall rank (Appendix C).

### 4.3 Graph-Text Decision-Making (RQ2)

We examine whether relations help Jev and related choice-based models, and whether combining relations with text yields additional benefit. Table 2 compares the matched input conditions for node classification on ogbn-arxiv and candidate-based retrieval on STaRK-Prime.

Table 2. Graph-text decision performance under matched input conditions for all fourteen configurations.

| Model / Configuration | ogbn-arxiv — Accuracy (%) | | | | | STaRK-Prime — Hit@1 (%) | | | | |
|---|---|---|---|---|---|---|---|---|---|---|
| | T | G | TG | BAG | A | T | G | TG | BAG | A* |
| Jev-1.13.0 | 55 | **61** | 58 | 57 | 51 | 48 | 47 | 49 | 46 | 42 |
| Decider-2B v11 | 35 | 45 | 42 | 38 | 24 | 20 | 17 | 18 | 21 | 20 |
| Kev-4B | 29 | **61** | 27 | 27 | 41 | 39 | 19 | 26 | 35 | 34 |
| Laya | U | U | U | U | U | U | U | U | U | U |
| Qwen3.5-0.8B | 32 | 34 | 26 | 24 | 33 | 9 | 9 | 9 | 7 | 8 |
| Qwen3.5-2B | 25 | 12 | 19 | 23 | 15 | 16 | 10 | 16 | 12 | 13 |
| Qwen3.5-4B generation | 21 | 59 | 30 | 19 | 42 | 36 | 17 | 24 | 28 | 26 |
| Qwen3.5-9B | 32 | 55 | 41 | 36 | 39 | 43 | 25 | 31 | 38 | 33 |
| Qwen3.5-27B | 46 | **61** | 58 | 56 | 49 | 50 | 44 | 46 | 43 | 41 |
| Qwen3.8-27B | 62 | 60 | 71 | 70 | 51 | 53 | 50 | 51 | 51 | 48 |
| Qwen2.5-72B-Instruct | 51 | 58 | 59 | 52 | 45 | 39 | 35 | 33 | 35 | 36 |
| Qwen3.5-4B scoring | U | U | U | U | U | U | U | U | U | U |
| GPT-5.4 | **73** | **61** | 73 | 72 | 48 | 60 | 51 | 53 | 56 | 56 |
| GPT-6-Astra | **73** | 60 | **76** | **73** | **55** | **69** | **69** | **67** | **73** | **71** |

Values are ogbn-arxiv accuracy and STaRK-Prime Hit@1 (%); higher is better. Each dataset uses 100 fixed targets or queries per condition. U denotes unsupported, not zero. A is the ogbn-arxiv shared-label anchor condition; A* is the STaRK-Prime relation-ablation comparator. Invalid outputs count as incorrect. Appendix D provides macro-F1, paired intervals and coverage. Each input-condition column is ranked independently within its dataset; higher is better. Within each stated comparison group, best values are bold and second-distinct values underlined, including ties; U and undefined values are excluded.

### 4.3.1 Node Classification

Jev benefits from relational input over label anchors alone, but this does not translate into a clear gain from adding explicit edges when text is already available. On ogbn-arxiv, its G-A difference is +10 percentage points with a paired 95% interval of [+3,+18], whereas TG-BAG is only +1 point [-4,+7]. These contrasts answer different questions: relations improve the anchor-only setting, but their additional contribution is less clear in the text-rich setting. The result is consistent with either overlapping evidence or limited use of the added edges; the present comparisons do not distinguish those explanations.

The Jev-like models reinforce the need to distinguish relational prediction from effective graph-text combination. Kev matches Jev's 61% accuracy under G but falls to 27% under TG, while Decider falls from 45% to 42%. Jev's own TG accuracy of 58% also remains below its G result. Joint input is therefore not a generally reliable enhancement for these choice-based models, and its effect differs substantially by configuration. Kev's large decrease establishes a performance loss under added target and context text, not a specific mechanism such as distraction or input-length sensitivity.

Jev's main-panel TG accuracy is 58%, while its fixed-forty-class macro-F1 is 24.81%. Overall accuracy is consequently insufficient to characterize class-level performance. However, the difference also reflects class coverage in this sample and the convention that classes absent from both truth and predictions receive zero F1; it cannot be attributed solely to uneven performance among observed classes. The separate 200-target study likewise finds a small TG-BAG difference of +3.0 points [-1.0,+7.0] (Appendices D.8 and F.2–F.3; cohort definitions in Table A.5).

Language-model baselines achieve higher joint-input accuracy in some configurations, but do not remove the distinction between joint performance and explicit-edge benefit. Qwen3.8-27B reaches 71% under TG while gaining only one point over BAG; Astra reaches 76% with a three-point TG-BAG difference whose interval includes zero. Conversely, Decider, Qwen3.5-4B generation and Qwen2.5-72B have positive TG-BAG intervals in the reported pointwise analysis. The evidence therefore supports configuration-specific gains, rather than either universal usefulness or universal irrelevance of explicit edges (Appendix D.3).

### 4.3.2 Candidate-Based Retrieval

Jev shows no clear additional relational benefit in the current STaRK-Prime candidate-selection setting. Its T, G and TG Hit@1 values are close at 48%, 47% and 49%; TG-BAG is +3 points [-6,+12] and G-A* is +5 points [-5,+14]. Both intervals include zero. This limits the evidence for an increment from relations within the answer-containing candidate sets, rather than establishing an absence of relational ability or a result about end-to-end retrieval.

The related models also lack a consistent joint-input advantage, but their responses to text differ across tasks. Kev's G-to-TG change is +7 points on STaRK-Prime, reversing its large negative change on ogbn-arxiv; nevertheless, its TG-BAG difference is negative at -9 points [-17,-2]. Adding text to a relational condition and adding edges to a text-rich condition should therefore not be conflated. Decider's TG result similarly does not exceed its T or BAG result. Laya's unsupported forty-choice interface is a separate limitation, not a zero retrieval score.

The baselines provide higher absolute retrieval performance without eliminating input dependence: GPT-5.4 performs best under T, whereas Astra performs best under BAG. No supported configuration has a strictly positive TG-BAG interval in this panel (Appendix D.6). Together with the ogbn-arxiv results, this shows that the usefulness of an input combination must be established for the particular model and task, rather than inferred from the availability of more information.

## 4.4 Sequential Decision-Making on Graphs (RQ3)

We examine whether Jev and related models can turn successive choices into high-quality graph solutions, and whether heuristic proposals improve decisions beyond the benefit of the proposal mechanism itself. Figure 3 compares direct construction and heuristic-proposal selection. Table 3 summarizes Jev's results against matched controls; Appendix E gives all configurations and fixed rules.

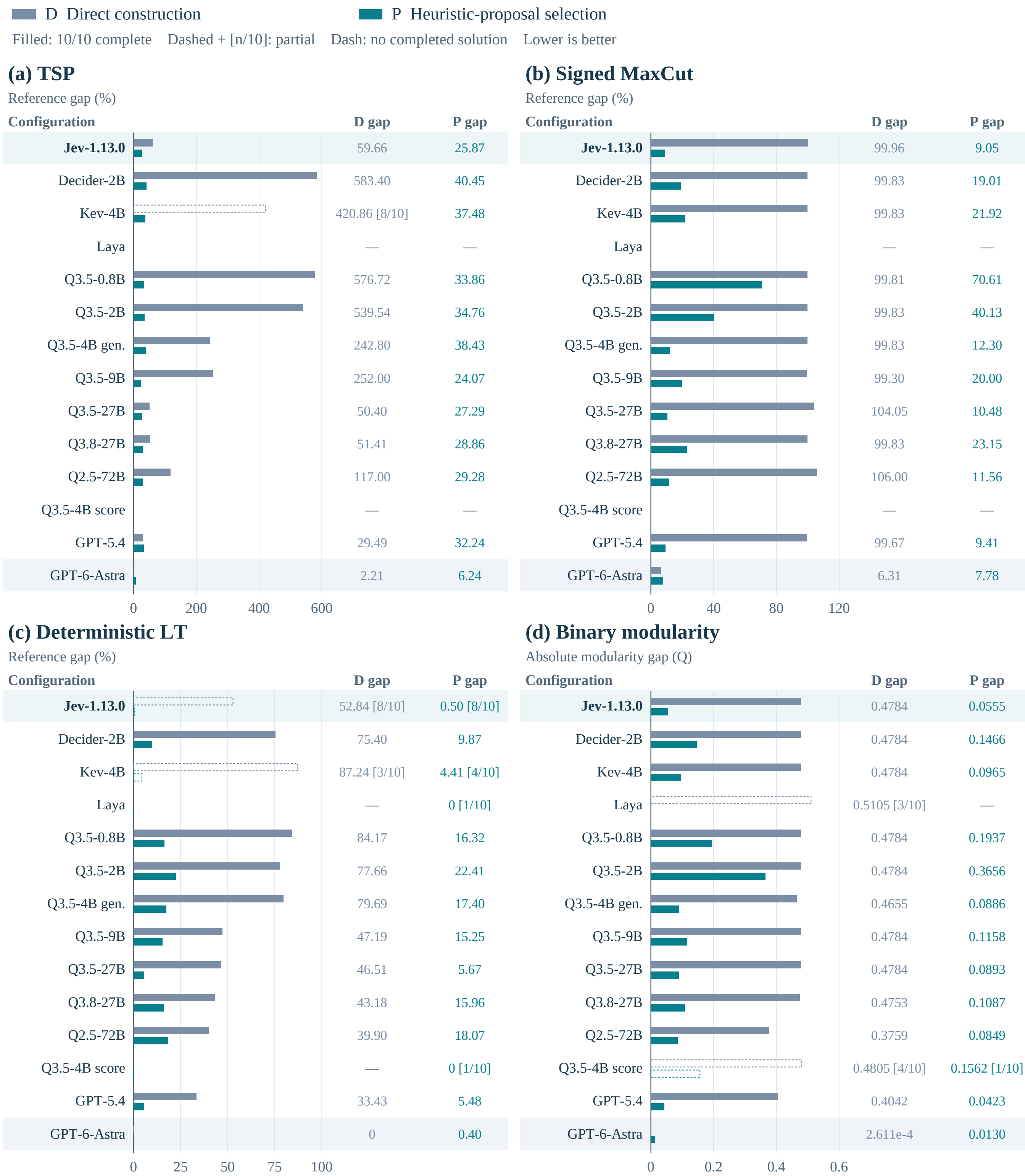

Figure 3. Mean reference gaps under direct construction (D) and heuristic-proposal selection (P); lower is better. All panels retain the same fourteen configurations. Filled bars denote 10/10 completed graphs; dashed outlines and [n/10] labels denote partial coverage. Numeric columns report both means, including small positive values. An exact zero uses a short vertical mark; a dash denotes no completed trajectory. TSP, MaxCut and LT use percent gaps; modularity uses absolute Q gaps. Axes are task-specific. Partial means are not directly rankable against full-coverage means. Forced-only completions do not establish model support; Astra is a reasoning-enabled reference. Appendix E gives matched comparisons.

Table 3. Jev solution gaps and matched controls; lower is better within each task.

| Task / units | Graphs | Direct | Proposal | Random proposal | Selected control |
|---|---|---|---|---|---|
| TSP (%) | 10/10 | 59.66 | 25.87 | 43.89 | return reserve<br>15.58 |
| Signed MaxCut (%) | 10/10 | 99.96 | 9.05 | 82.72 | greedy completion<br>8.87 |
| LT (%) | 8/10 | 52.84 | 0.50 | 21.91 | classical<br>6.84 |
| Binary modularity (Q) | 10/10 | 0.4784 | 0.0555 | 0.4175 | greedy completion<br>0.0361 |

All columns within a row use the same completed graphs. The LT row uses the eight graphs completed in both modes, not the full ten-graph control mean. The other rows use all ten graphs. Selected controls illustrate task-specific comparisons, not a single universal baseline. Appendix E lists every fixed rule and classical control.

### 4.4.1 Direct Construction

Jev can complete construction without producing a strong solution. It finishes all TSP, MaxCut and modularity instances, yet its mean TSP gap is 59.66%, its mean MaxCut objective is zero, and its mean modularity is also zero. Its completed LT instances have a 52.84% gap. These weaknesses span many-option route construction, binary partition decisions and a two-seed selection problem, so neither a large action space nor a long decision sequence alone explains the pattern. The common challenge is making local choices that serve the final objective, not simply identifying legal next actions.

Task-specific controls show the practical size of the shortfall. Jev's TSP gap remains above the 43.89% within-proposal-pool random control and the 5.75% classical control, although the former already benefits from heuristic candidate construction and is not a full-action random policy. On signed MaxCut, positive and negative attained objectives average to zero; this does not mean every graph receives the same partition or objective. Its 99.96% mean gap contrasts with 19.73% for assigned-prefix gain. For modularity, mean Q=0 matches the single-community baseline in objective value, but does not by itself identify the produced partitions. These comparisons expose weak optimization despite successful completion.

The Jev-like results separate execution coverage from decision quality. Decider completes all eighty evaluation units across both modes, but its direct TSP gap is 583.40% and its direct LT gap is 75.40%. Kev also constructs weak direct solutions and has task-dependent support limitations. Jev's LT means cover only eight graphs because both modes fail on the same two instances; on its completed subset, the matched classical gap is 6.84%, well below its direct gap. Across all configurations, the 1,120 scheduled evaluation units comprise 951 feasible completions, 165 unsupported units and four failures, all in Jev LT. Appendix B records their terminal causes and coverage.

The language-model baselines confirm that the direct-construction interface permits much stronger policies, but model scale alone is not sufficient. Larger Qwen models remain weak on several tasks, especially signed MaxCut, whereas Astra obtains much smaller gaps across all four tasks and reaches the exact LT references. These baselines locate the limitations of the evaluated System One configurations; differences between distinct models and reasoning settings do not isolate the causal contribution of reasoning.

### 4.4.2 Heuristic-Proposal Selection

Heuristic proposals consistently improve Jev on the instances completed under both modes: all ten TSP, ten MaxCut, eight LT and ten modularity instances have lower gaps under proposal selection. This is useful assistance under the evaluated configuration, but the intervention supplies candidate actions, numerical features, rule identities and sometimes forced singleton transitions together. Its benefit cannot be attributed solely to reducing the number of options, particularly in the already binary partition tasks.

For Jev, beating its direct policy does not necessarily mean beating a fixed rule. Its TSP proposal gap of 25.87% improves on random selection within the pool, but is almost identical to the nearest rule's 25.86% and remains above return reserve's 15.58%. Its MaxCut gap of 9.05% is far better than the 82.72% random-proposal gap, yet slightly worse than greedy completion's 8.87%. Similarly, its modularity gap falls to 0.0555 but remains above greedy completion's 0.0361. These results show useful selection relative to random proposals without a consistent advantage over the strongest fixed rule.

LT is a stronger case for Jev's assisted performance. On the same eight completed graphs, its proposal gap is 0.50%, below both the 21.91% random-candidate gap and the 6.84% classical gap. This supports effective use of the supplied proposals on that subset, not a comparison against another model's full ten-graph mean. One of these completions requires no model call because both seed choices are forced; the zero-gap LT completions of Laya and candidate scoring are also zero-call episodes and do not demonstrate successful model selection.

The Jev-like models also benefit from proposals, although their assisted quality and coverage differ. Decider's large improvements from weak direct policies still leave it behind Jev on the same ten-graph TSP, MaxCut and modularity panels. Kev's TSP direct mode completes eight of ten instances, with two unsupported; its proposal mode completes all ten. The [8/10] label in Figure 3 therefore marks a conditional mean, not 80% accuracy. On the eight graphs completed in both modes, its mean gap improves by 384.23 percentage points (Appendix E.1). This paired result, rather than subtraction of means over different subsets, establishes the improvement.

The stronger baselines bound the generality of assistance. GPT-5.4 improves in mean gap on MaxCut, LT and modularity, but not TSP; Astra's mean gap increases on each of the four tasks. The latter is an aggregate statement, not deterioration on every instance: its LT gap increases from 0 to 0.4% because one instance worsens while nine remain tied. Proposal restrictions may exclude useful actions, but features and decision histories also change, so the observed losses do not identify a unique mechanism.

Overall, Jev benefits substantially from heuristic guidance, while surpassing a single fixed rule remains task-dependent. Related choice-based models show the same need to distinguish an improvement over a weak direct policy from strong final optimization performance. Comparisons against direct construction, random selection within the proposal pool and fixed rules reveal different levels of value; none can replace the others.

# 5 Discussion and Conclusion

## 5.1 Main Findings and Implications

**Structural competence is task-specific, even within a common choice interface.** Jev combines strong local connection recognition with weaker exact counting and articulation judgments. The different profiles of Decider, Kev and Laya show that related choice-based interfaces do not imply interchangeable graph capabilities. For applications, successful edge recognition is therefore insufficient evidence that a model can reliably carry out other structural operations. The relevant capability must be assessed at the operation required by the task, rather than inferred from the interface or a pooled score.

**Useful relational input does not guarantee effective graph-text combination.** Jev benefits from relations over label anchors on ogbn-arxiv, but the additional benefit of explicit edges is less clear when text is already supplied. The Jev-like models respond differently to joint input, and the STaRK-Prime results do not reproduce the same pattern of relational gains. Input design should thus consider what additional evidence a model can use, not simply how much information can be included. Matched contrasts assess this incremental value without treating high joint-input accuracy as evidence of effective use of every input component.

**Heuristic guidance is useful, but its value must be distinguished from adaptive selection.** Proposal assistance substantially improves Jev's construction results, while surpassing a fixed rule remains task-dependent. Related models also improve from weak direct policies without uniformly attaining strong solutions. Random selection within the proposal pool and fixed-rule controls help assess the model's additional value over the supplied guidance. Worse assisted results for stronger baselines further show that assistance is not universally preferable: candidate restrictions are one possible explanation, but changes in features and decision histories prevent a unique attribution.

## 5.2 Scope and Limitations

The benchmark covers limited public instances, small graph-cognition inputs and specific construction tasks; the findings do not establish performance on larger or dynamic graphs. In ogbn-arxiv, shared context and label anchors induce dependence between targets. STaRK-Prime evaluates selection within answer-containing candidate sets rather than end-to-end retrieval. The supplementary graph-text evaluation uses a separate target cohort and model settings; its results are not pooled with the primary estimates.

Comparisons concern model-interface configurations and do not isolate effects of architecture, scale or reasoning. Proposal assistance changes candidate actions, features, forced transitions and subsequent histories together, so its mechanisms are not separately identified. Classical controls may also revise solutions, unlike irreversible direct construction. Quality comparisons with partial completion require matched instance subsets. Single executions do not estimate within-instance stochastic variation, and paired target-bootstrap intervals neither replace repeated-run uncertainty estimates nor fully account for shared graph context. The reported intervals are pointwise and unadjusted for multiple comparisons.

### 5.3 Conclusion

GraphDecisionBench provides a model-independent benchmark for studying System One models through graph-structural judgments, graph-text information use and sequential solution construction. Jev exhibits competitive but uneven structural capabilities, condition-dependent benefits from relational input, and substantial improvements from heuristic proposals without consistently surpassing fixed rules. The differing profiles of related choice-based models show that these capabilities cannot be inferred from a shared selection interface alone. The results characterize both the promise and the limitations of the evaluated System One configurations and support assessing their usefulness under the task, information and candidate conditions in which they will be applied. Separating these diagnostic comparisons from model-specific adapters makes the benchmark applicable to future decision models as well as the configurations studied here.

# Appendix A. Data and Task Specifications

Appendices A and B specify tasks, data, cohorts and decision interfaces; C–E report the primary results and controls, with the separate graph-text cohort identified in D.8. Appendix F adds interface details and supplementary graph-text evaluation, and G documents optimization reference calculations. Primary and supplementary graph-text cohorts retain separate denominators and are not pooled.

## A.1 Structural Queries

Let G = (V, E) be the supplied simple, undirected, unweighted induced graph, with n = |V|. The explicit vertex set includes isolated vertices. Every query concerns this supplied graph, not the full source network. The six task contracts are:

- **Adjacency.** Given distinct vertices u and v, determine whether the undirected edge {u, v} belongs to E.
- **Exact degree.** Given vertex v, return the number of its neighbors. The answer options are all integers from 0 to n-1.
- **Cycle existence.** Determine whether any connected component contains a simple cycle through at least three distinct vertices. Traversing one undirected edge back and forth is not a cycle.
- **Pairwise connectivity.** Given distinct vertices u and v, determine whether an undirected path connects them. Other components may be disconnected: this is not a test of whole-graph connectivity.
- **Distance threshold.** Given distinct vertices u and v and threshold k, determine whether the unweighted shortest-path length $d_G(u, v) \le k$. The primary evaluation uses k = 2. An unreachable pair has infinite distance and therefore receives a no answer.
- **Articulation point.** Given vertex v, determine whether deleting v and its incident edges increases the number of connected components relative to the original graph. The original graph need not be connected.

All tasks except exact degree supply yes/no options. Independent graph algorithms compute the reference answers. For each of the four sources and each size n = 8, 9, 10, 11, 12, every binary task contains two yes and two no queries, and the degree task contains two queries for each possible degree. Thus degree contributes 400 queries and each other task contributes 80. Sampling is answer-conditioned: degree queries select vertices to meet the prescribed degree, while binary queries select qualifying induced samples and query vertices without editing graph topology. These balanced quotas do not reproduce the natural label frequencies of the source networks. Full results appear in Appendix C.

## A.2 Sequential Objectives and Feasibility

TSP starts at the fixed initial city, appends unvisited cities to a partial route, and visits every city once before closing the tour. The last unvisited city and the return edge are completed automatically. EUC_2D distances round the Euclidean distance to the nearest integer as specified by TSPLIB. Signed MaxCut assigns each vertex to one of two sides and maximizes the sum of signed weights on crossing edges; negative weights remain negative. Partition construction fixes the next vertex rather than allowing a vertex-order search. The current state and partial solution are carried forward after each action.

LT chooses two distinct seeds before diffusion starts and activates them simultaneously. Updates are synchronous and irreversible. Each incoming edge has weight 1/indegree and every nonseed has threshold 1/2; equality activates a vertex, and a zero-indegree nonseed never activates. The objective is the final number of active vertices. Exhaustive evaluation of all unordered seed pairs supplies the exact reference. This fixed-threshold process is not the random-threshold model underlying the usual influence-maximization approximation guarantee [9].

The lack of a diminishing-returns guarantee follows even in a small deterministic example: let a, b and d each point to c with weight 1/3. Vertex c requires two active parents. Spread is zero for the empty set, one for either seed a or b alone, and three for seeds {a,b}. Adding b therefore gains one at the empty set but two at {a}. This task property motivates evaluating marginal-cascade heuristics empirically rather than transferring the random-threshold guarantee.

**Binary modularity.** For the supplied undirected graph, let $A_{uv}$ be its adjacency-matrix entry, $d_u$ the degree of u, m = |E| the number of edges, and $c_v$ the community label of v. The indicator equals one when the two labels agree and zero otherwise. The sum is over ordered vertex pairs:

$$Q(c) = \frac{1}{2m} \sum_{u,\, v \in V} \left( A_{uv} - \frac{d_u d_v}{2m} \right) \mathbf{1}[c_u = c_v], \qquad c_v \in \{0, 1\}. \tag{A.1}$$

Binary modularity uses full-graph degrees, resolution one, and at most two communities [38]. A single-community assignment is legal and gives Q=0. Results include attained Q and gaps to the numerical solver-certified optima in Appendix G.1.

### A.3 Public Optimization Instances

The following inventory gives every RQ3 graph, size and edge count. TSP references are proven optima from TSPLIB [22]; MaxCut references are historical 2016 best-known values from the OPTSICOM collection [5, 17]; LT references are exact seed-pair optima. The modularity graph inventory is supplemented by the certified reference values in Table G.1. Each graph contributes one condition per arm. LT and binary modularity each include two induced subgraphs at each size 50, 75, 100, 125 and 150, indexed s0 and s1. Appendix G.1 gives the binary-modularity MILP and numerical certification; Appendix G.2 gives deterministic LT seed-pair enumeration and independent closure checks. Both sections list all ten fixed-instance references.

Table A.1. TSP: proven optimum.

| Instance | n | Edges | Reference |
|---|---|---|---|
| kroA100 | 100 | 4950 | 21282 |
| kroA150 | 150 | 11175 | 26524 |
| kroB100 | 100 | 4950 | 22141 |
| kroB200 | 200 | 19900 | 29437 |
| kroC100 | 100 | 4950 | 20749 |
| bier127 | 127 | 8001 | 118282 |
| tsp225 | 225 | 25200 | 3916 |
| eil76 | 76 | 2850 | 538 |
| rat99 | 99 | 4851 | 1211 |
| kroD100 | 100 | 4950 | 21294 |

Table A.2. Signed MaxCut: 2016 best-known.

| Instance | n | Edges | Reference |
|---|---|---|---|
| sg3dl051000 | 125 | 375 | 110 |
| sg3dl052000 | 125 | 375 | 112 |
| sg3dl053000 | 125 | 375 | 106 |
| sg3dl054000 | 125 | 375 | 114 |
| sg3dl055000 | 125 | 375 | 112 |
| sg3dl056000 | 125 | 375 | 110 |
| sg3dl057000 | 125 | 375 | 112 |
| sg3dl058000 | 125 | 375 | 108 |
| sg3dl059000 | 125 | 375 | 110 |
| sg3dl0510000 | 125 | 375 | 112 |

Table A.3. Deterministic LT: exact seed-pair optimum.

| Instance | n | Edges | Reference |
|---|---|---|---|
| email:n50-s0 | 50 | 150 | 14 |
| email:n50-s1 | 50 | 128 | 17 |
| email:n75-s0 | 75 | 317 | 29 |
| email:n75-s1 | 75 | 232 | 63 |
| email:n100-s0 | 100 | 348 | 25 |
| email:n100-s1 | 100 | 470 | 25 |
| email:n125-s0 | 125 | 507 | 25 |
| email:n125-s1 | 125 | 643 | 16 |
| email:n150-s0 | 150 | 843 | 17 |
| email:n150-s1 | 150 | 1002 | 15 |

Table A.4. Binary-modularity graph inventory.

| Instance | n | Edges |
|---|---|---|
| GrQc:n50-s0 | 50 | 85 |
| GrQc:n50-s1 | 50 | 69 |
| GrQc:n75-s0 | 75 | 92 |
| GrQc:n75-s1 | 75 | 139 |
| GrQc:n100-s0 | 100 | 226 |
| GrQc:n100-s1 | 100 | 180 |
| GrQc:n125-s0 | 125 | 231 |
| GrQc:n125-s1 | 125 | 199 |
| GrQc:n150-s0 | 150 | 204 |
| GrQc:n150-s1 | 150 | 224 |

### A.4 Evaluation Cohorts

Table A.5 lists four cohorts: primary RQ1, primary RQ2, primary RQ3 and supplementary graph-text evaluation on 200 ogbn-arxiv targets. Each cohort retains its own graph, target or query set, model-interface settings and evaluation denominator. The five graph-text conditions are matched views of the same targets or queries, not additional independent samples, and results from different cohorts are not pooled.

Table A.5. Primary and supplementary evaluation cohorts.

| Cohort | Sampling unit / size | Comparison | Protocol / results |
|---|---|---|---|
| Primary RQ1 | 800 queries; four sources; five sizes | Six structural tasks | Appendices A.1, C |
| Primary RQ2 | 100 targets + 100 queries | Five conditions per dataset; 1,000 decisions/configuration | Sections 3.2, 4.3; Appendix D.1–D.7 |
| Primary RQ3 | Ten graphs per task; four tasks | Two modes; 80 trajectories/configuration | Section 4.4; Appendices A, B, E |
| Supplementary graph-text | 200 ogbn-arxiv targets | Five conditions; separate model settings | Appendices D.8, F.2–F.3 |

## Appendix B. Interfaces and Candidate Rules

### B.1 Model Configurations

Table B.1. Served models, interfaces, and configured output values.

| Configuration | Provider / interface | Served model | RQ1 / RQ3 | RQ2 |
|---|---|---|---|---|
| Jev-1.13.0 | typesafe | jev-1.13.0 | -- | -- |
| Decider-2B v11 | plugin | Mapika/decider-2b | -- | -- |
| Kev-4B | plugin | jaredpalmer/kev-4b | -- | -- |
| Laya | plugin | convaiinnovations/laya-typed-decisions | -- | -- |
| Qwen3.5-0.8B | vllm | Qwen3.5-0.8B | 64 | 16 |
| Qwen3.5-2B | vllm | Qwen3.5-2B | 64 | 16 |
| Qwen3.5-4B generation | vllm | Qwen3.5-4B | 64 | 16 |
| Qwen3.5-9B | vllm | Qwen3.5-9B | 64 | 16 |
| Qwen3.5-27B | vllm | Qwen3.5-27B | 64 | 16 |
| Qwen3.8-27B | vllm | Qwen3.8-27B | 64 | 16 |
| Qwen2.5-72B-Instruct | vllm | Qwen2.5-72B | 64 | 16 |
| Qwen3.5-4B scoring | plugin | Qwen3.5-4B | -- | -- |
| GPT-5.4 | github_copilot | gpt-5.4 | 4096 | 4096 |
| GPT-6-Astra | github_copilot | gpt-6-astra | 4096 | 4096 |

Q3.5/Q3.8/Q2.5 abbreviate Qwen3.5/Qwen3.8/Qwen2.5 in result tables. gen. and score distinguish generation and scoring interfaces. A dash means no value is specified for that interface.

Qwen generation uses no-thinking mode with 64 output tokens for RQ1/RQ3 and 16 for RQ2. Native/scoring interfaces retain their support constraints. GPT-5.4 uses provider github_copilot, served model gpt-5.4, think=False, reasoning_effort="none", output_format=function_call and max_tokens=4096, with answer_only task instructions for RQ2. Temperature, model seed, top_p, top_k, presence_penalty and constrain_choices are unset. Astra leaves think and reasoning_effort unset in the evaluated request configuration; medium is observed in response metadata rather than explicitly set in the request. Other unspecified settings retain provider defaults. The configured output value of 4,096 is not a verified limit on total reasoning tokens.

Thirty LT trajectories from Qwen3.5-27B, Qwen3.8-27B and Qwen2.5-72B use 40,960-token contexts. Ten of Qwen2.5-72B's LT trajectories additionally use static YaRN factor 4 with original_max_position_embeddings=32768;

four of its TSP direct-construction trajectories use BF16, a 40,960-token context and the same YaRN setting. Other trajectories retain their respective context settings.

### B.2 Fixed Candidate Rules and Controls

RQ2 classification supplies all forty classes, and STaRK-Prime supplies forty distinct candidate entities. For RQ3 direct construction, the legal actions are unvisited cities for TSP, unselected vertices for seed selection, and either side for the current vertex in MaxCut and modularity. At each state, Heuristic-proposal selection merges duplicate actions nominated by four fixed rules and retains their rule identities. Only one action is executed. A singleton is forced and incurs no model call. The random-candidate control selects uniformly among distinct candidates, averaging five seeds within each graph before averaging across graphs. Each fixed-rule control follows that rule throughout construction; the reported classical control is listed separately.

**TSP.** For current city c, start s, and unvisited set U, let e(v) be the distance from c to v, r(v) the distance from v to s, and h(v) the distance from v to its nearest other unvisited city. The rules nearest, one_step_lookahead, return_reserve, and isolation_priority minimize e, e+h, 2e-r, and e-h, respectively. Rules propose a next city, not a repaired or locally optimized tour. Ties use the lowest vertex identifier.

**Partition tasks.** The four rules are assigned_gain, greedy_completion, opposite_completion, and degree_balance. They evaluate assigned-prefix gain, full greedy completion, completion with the remaining suffix on the opposite side, and degree-volume balance, respectively. Cut gains preserve signed weights, while cut volume uses absolute incident weights. For modularity, with current degree d, m full-graph edges, e_s edges to assigned side s, and pre-action volume V_s, assigned gain is proportional to $4me_s-2dV_s-d^2$. Completion estimates score a full partition but execute only the next vertex assignment. Because Direct construction already has two sides, Heuristic-proposal selection is either the same two actions or a forced singleton.

**LT.** The rules marginal_cascade, outdegree, two_hop_reach, and threshold_deficit maximize marginal deterministic spread, outdegree, inactive two-hop reach, and threshold relief. With active closure X, threshold relief is zero for v in X. For an inactive v, it sums reciprocal residual activation deficits over its inactive out-neighbors; the deficit at t is ceil(indegree(t)/2) minus its active in-neighbor count. The two selected seeds are activated together for final scoring. These task-specific rules are fixed rather than learned.

**STaRK-Prime candidate construction.** Candidate construction starts with BM25 top-40. If no native correct answer is present, the lowest-ranked candidate is replaced by a deterministically selected correct entity; the final set is deterministically shuffled and its text and relational evidence rebuilt without answer markers. Candidate Hit@40 is therefore 100% by design, while candidate Recall@40 is 77.7241% because a multi-answer query need not retain every correct entity.

### B.3 Scoring, Statistics, and Coverage

RQ2 paired intervals use 1,000 target-bootstrap resamples, seed 20261001 and sorted zero-based endpoints 24 and 974. Conditions of a resampled target stay paired. For GPT-5.4, numeric query IDs are sorted and Python random.Random draws indices uniformly with replacement, using common samples across contrasts. These are pointwise 95% intervals without multiplicity adjustment. Fixed-forty-class macro-F1 gives zero F1 to classes absent from both truth and predictions. Shared ogbn-arxiv contexts and anchors induce dependence beyond target resampling. RQ3 means condition on feasible completion, while coverage retains every scheduled trajectory without objective imputation.

Table B.2. Primary RQ3 coverage: eighty scheduled evaluation units per configuration.

| Configuration | Complete | Unsupported | Failed | Coverage |
|---|---|---|---|---|
| Jev-1.13.0 | 76 | 0 | 4 | 95.00% |
| Decider-2B | 80 | 0 | 0 | **100.00%** |
| Kev-4B | 65 | 15 | 0 | 81.25% |
| Laya | 4 | 76 | 0 | 5.00% |
| Q3.5-0.8B | 80 | 0 | 0 | **100.00%** |
| Q3.5-2B | 80 | 0 | 0 | **100.00%** |
| Q3.5-4B gen. | 80 | 0 | 0 | **100.00%** |
| Q3.5-9B | 80 | 0 | 0 | **100.00%** |
| Q3.5-27B | 80 | 0 | 0 | **100.00%** |
| Q3.8-27B | 80 | 0 | 0 | **100.00%** |
| Q2.5-72B | 80 | 0 | 0 | **100.00%** |
| Q3.5-4B score | 6 | 74 | 0 | 7.50% |
| GPT-5.4 | 80 | 0 | 0 | **100.00%** |
| GPT-6-Astra | 80 | 0 | 0 | **100.00%** |

Coverage alone is ranked as a completion rate over the same 80 scheduled evaluation units, not as solution quality.

RQ1 has 11,200 legal outputs. Supported RQ2 units have 11,998 legal and two illegal outputs, the latter both GPT-5.4 STaRK-Prime TG responses; illegal outputs count as incorrect. The two unsupported configurations remain explicit. The four RQ3 failures are Jev LT http_error terminals caused by max_tokens_exceeded, listed below. Unsupported trajectories are counted separately. Forced-singleton completion may require no model call, so the few Laya/scoring completions do not imply general interface support.

Table B.3. Retained primary RQ3 failures.

| Configuration | Graph | Mode | Terminal | Cause |
|---|---|---|---|---|
| Jev-1.13.0 | email:n150-s1 | Heuristic-proposal selection | http_error | max_tokens_exceeded |
| Jev-1.13.0 | email:n150-s1 | Direct construction | http_error | max_tokens_exceeded |
| Jev-1.13.0 | email:n150-s0 | Heuristic-proposal selection | http_error | max_tokens_exceeded |
| Jev-1.13.0 | email:n150-s0 | Direct construction | http_error | max_tokens_exceeded |

## B.4 Complete GPT-5.4 RQ3 Results

Tables B.4a–B.4h report all eighty GPT-5.4 no-reasoning trajectories, ten graphs for each task and construction mode. Every row is feasible with legal output; objectives and gaps are rounded only for display. TSP, MaxCut and LT gaps are percentages; modularity uses absolute Q*-Q. A zero-call LT proposal trajectory follows forced singleton actions. Task summaries and matched controls appear in Appendix E.

Table B.4a. GPT-5.4: TSP, Direct construction.

| Graph | Objective | Gap (%) | Calls |
|---|---|---|---|
| bier127 | 159001 | 34.425356 | 125 |
| eil76 | 764 | 42.007435 | 74 |
| kroA100 | 27515 | 29.287661 | 98 |
| kroA150 | 34161 | 28.792791 | 148 |
| kroB100 | 28231 | 27.505533 | 98 |
| kroB200 | 39284 | 33.451099 | 198 |
| kroC100 | 26332 | 26.907321 | 98 |
| kroD100 | 26998 | 26.786888 | 98 |
| rat99 | 1401 | 15.689513 | 97 |
| tsp225 | 5091 | 30.005107 | 223 |

Table B.4b. GPT-5.4: TSP, Heuristic-proposal selection.

| Graph | Objective | Gap (%) | Calls |
|---|---|---|---|
| bier127 | 157666 | 33.296698 | 69 |
| eil76 | 682 | 26.765799 | 39 |
| kroA100 | 28896 | 35.776713 | 53 |
| kroA150 | 36107 | 36.129543 | 86 |
| kroB100 | 28478 | 28.621110 | 52 |
| kroB200 | 37028 | 25.787275 | 101 |
| kroC100 | 26819 | 29.254422 | 44 |
| kroD100 | 27337 | 28.378886 | 41 |
| rat99 | 1636 | 35.094963 | 57 |
| tsp225 | 5613 | 43.335036 | 126 |

Table B.4c. GPT-5.4: Signed MaxCut, Direct construction.

| Graph | Objective | Gap (%) | Calls |
|---|---|---|---|
| sg3dl051000 | -6 | 105.454545 | 124 |
| sg3dl0510000 | 8 | 92.857143 | 124 |
| sg3dl052000 | -6 | 105.357143 | 124 |
| sg3dl053000 | -10 | 109.433962 | 124 |
| sg3dl054000 | -2 | 101.754386 | 124 |
| sg3dl055000 | -6 | 105.357143 | 124 |
| sg3dl056000 | 26 | 76.363636 | 124 |
| sg3dl057000 | 10 | 91.071429 | 124 |
| sg3dl058000 | 4 | 96.296296 | 124 |
| sg3dl059000 | -14 | 112.727273 | 124 |

Table B.4d. GPT-5.4: Signed MaxCut, Heuristic-proposal selection.

| Graph | Objective | Gap (%) | Calls |
|---|---|---|---|
| sg3dl051000 | 96 | 12.727273 | 90 |
| sg3dl0510000 | 104 | 7.142857 | 86 |
| sg3dl052000 | 96 | 14.285714 | 105 |
| sg3dl053000 | 96 | 9.433962 | 118 |
| sg3dl054000 | 108 | 5.263158 | 116 |
| sg3dl055000 | 106 | 5.357143 | 84 |
| sg3dl056000 | 94 | 14.545455 | 94 |
| sg3dl057000 | 100 | 10.714286 | 104 |
| sg3dl058000 | 102 | 5.555556 | 96 |
| sg3dl059000 | 100 | 9.090909 | 94 |

Table B.4e. GPT-5.4: Deterministic LT, Direct construction.

| Graph | Objective | Gap (%) | Calls |
|---|---|---|---|
| email:n100-s0 | 7 | 72.000000 | 2 |
| email:n100-s1 | 15 | 40.000000 | 2 |
| email:n125-s0 | 12 | 52.000000 | 2 |
| email:n125-s1 | 7 | 56.250000 | 2 |
| email:n150-s0 | 10 | 41.176471 | 2 |
| email:n150-s1 | 11 | 26.666667 | 2 |
| email:n50-s0 | 10 | 28.571429 | 2 |
| email:n50-s1 | 14 | 17.647059 | 2 |
| email:n75-s0 | 29 | 0.000000 | 2 |
| email:n75-s1 | 63 | 0.000000 | 2 |

Table B.4f. GPT-5.4: Deterministic LT, Heuristic-proposal selection.

| Graph | Objective | Gap (%) | Calls |
|---|---|---|---|
| email:n100-s0 | 17 | 32.000000 | 2 |
| email:n100-s1 | 24 | 4.000000 | 1 |
| email:n125-s0 | 25 | 0.000000 | 1 |
| email:n125-s1 | 13 | 18.750000 | 1 |
| email:n150-s0 | 17 | 0.000000 | 1 |
| email:n150-s1 | 15 | 0.000000 | 1 |
| email:n50-s0 | 14 | 0.000000 | 2 |
| email:n50-s1 | 17 | 0.000000 | 2 |
| email:n75-s0 | 29 | 0.000000 | 0 |
| email:n75-s1 | 63 | 0.000000 | 1 |

Table B.4g. GPT-5.4: Binary modularity, Direct construction.

| Graph | Objective | Gap (Q) | Calls |
|---|---|---|---|
| GrQc:n100-s0 | 0.1101006343 | 0.3771340747 | 99 |
| GrQc:n100-s1 | 0.0647993827 | 0.3907407407 | 99 |
| GrQc:n125-s0 | 0.0015367028 | 0.4531680441 | 124 |
| GrQc:n125-s1 | 0.1629251787 | 0.3270119441 | 124 |
| GrQc:n150-s0 | 0.0564686659 | 0.4313725490 | 149 |
| GrQc:n150-s1 | 0.1579141422 | 0.3286830357 | 149 |
| GrQc:n50-s0 | 0.0481660900 | 0.4400692042 | 49 |
| GrQc:n50-s1 | 0.1797941609 | 0.3019323671 | 49 |
| GrQc:n75-s0 | 0.0145912098 | 0.4673913043 | 74 |
| GrQc:n75-s1 | -0.0548625848 | 0.5248951918 | 74 |

Table B.4h. GPT-5.4: Binary modularity, Heuristic-proposal selection.

| Graph | Objective | Gap (Q) | Calls |
|---|---|---|---|
| GrQc:n100-s0 | 0.4831917143 | 0.0040429948 | 84 |
| GrQc:n100-s1 | 0.4209722222 | 0.0345679012 | 50 |
| GrQc:n125-s0 | 0.4460467383 | 0.0086580087 | 80 |
| GrQc:n125-s1 | 0.4045226131 | 0.0854145097 | 100 |
| GrQc:n150-s0 | 0.4117166474 | 0.0761245675 | 109 |
| GrQc:n150-s1 | 0.4329460300 | 0.0536511480 | 98 |
| GrQc:n50-s0 | 0.4210380623 | 0.0671972318 | 22 |
| GrQc:n50-s1 | 0.4683889939 | 0.0133375341 | 39 |
| GrQc:n75-s0 | 0.4021148393 | 0.0798676749 | 55 |
| GrQc:n75-s1 | 0.4700326070 | 0.0000000000 | 67 |

## B.5 Selected Episode Outcomes

Table B.5 lists selected primary-evaluation episodes, including context-expanded LT inputs and Jev's input rejection. These rows also contribute to the aggregate results.

Table B.5. Selected episode outcomes: terminal status and confirmed calls.

| Configuration | Graph | Mode | Feasible | Terminal | Calls |
|---|---|---|---|---|---|
| Qwen3.5-27B | email-Eu-core-main-n75-s1 | Direct construction | True | complete | 2 |
| Qwen3.5-27B | sg3dl055000 | Direct construction | True | complete | 124 |
| Qwen3.5-27B | sg3dl0510000 | Heuristic-proposal selection | True | complete | 92 |
| Jev-1.13.0 | email-Eu-core-main-n150-s1 | Heuristic-proposal selection | False | http_error | 1 |
| Qwen2.5-72B | email-Eu-core-main-n100-s1 | Heuristic-proposal selection | True | complete | 2 |
| Qwen2.5-72B | email-Eu-core-main-n150-s1 | Heuristic-proposal selection | True | complete | 1 |
| Qwen2.5-72B | email-Eu-core-main-n125-s1 | Heuristic-proposal selection | True | complete | 2 |
| Qwen2.5-72B | email-Eu-core-main-n100-s1 | Direct construction | True | complete | 2 |
| Qwen2.5-72B | email-Eu-core-main-n150-s1 | Direct construction | True | complete | 2 |
| Qwen2.5-72B | email-Eu-core-main-n125-s1 | Direct construction | True | complete | 2 |
| Qwen2.5-72B | email-Eu-core-main-n150-s0 | Heuristic-proposal selection | True | complete | 1 |

| Configuration | Graph | Mode | Feasible | Terminal | Calls |
|---|---|---|---|---|---|
| Qwen2.5-72B | email-Eu-core-main-n125-s0 | Heuristic-proposal selection | True | complete | 1 |
| Qwen2.5-72B | email-Eu-core-main-n125-s0 | Direct construction | True | complete | 2 |
| Qwen2.5-72B | email-Eu-core-main-n150-s0 | Direct construction | True | complete | 2 |
| Qwen3.5-27B | email-Eu-core-main-n100-s1 | Heuristic-proposal selection | True | complete | 2 |
| Qwen3.5-27B | email-Eu-core-main-n150-s1 | Heuristic-proposal selection | True | complete | 1 |
| Qwen3.5-27B | email-Eu-core-main-n125-s1 | Heuristic-proposal selection | True | complete | 2 |
| Qwen3.5-27B | email-Eu-core-main-n100-s1 | Direct construction | True | complete | 2 |
| Qwen3.5-27B | email-Eu-core-main-n150-s1 | Direct construction | True | complete | 2 |
| Qwen3.5-27B | email-Eu-core-main-n125-s1 | Direct construction | True | complete | 2 |
| Qwen3.5-27B | email-Eu-core-main-n150-s0 | Heuristic-proposal selection | True | complete | 1 |
| Qwen3.5-27B | email-Eu-core-main-n125-s0 | Heuristic-proposal selection | True | complete | 1 |
| Qwen3.5-27B | email-Eu-core-main-n125-s0 | Direct construction | True | complete | 2 |
| Qwen3.5-27B | email-Eu-core-main-n150-s0 | Direct construction | True | complete | 2 |
| Qwen3.8-27B | email-Eu-core-main-n100-s1 | Heuristic-proposal selection | True | complete | 2 |
| Qwen3.8-27B | email-Eu-core-main-n150-s1 | Heuristic-proposal selection | True | complete | 1 |
| Qwen3.8-27B | email-Eu-core-main-n125-s1 | Heuristic-proposal selection | True | complete | 1 |
| Qwen3.8-27B | email-Eu-core-main-n100-s1 | Direct construction | True | complete | 2 |
| Qwen3.8-27B | email-Eu-core-main-n150-s1 | Direct construction | True | complete | 2 |
| Qwen3.8-27B | email-Eu-core-main-n125-s1 | Direct construction | True | complete | 2 |
| Qwen3.8-27B | email-Eu-core-main-n150-s0 | Heuristic-proposal selection | True | complete | 1 |
| Qwen3.8-27B | email-Eu-core-main-n125-s0 | Heuristic-proposal selection | True | complete | 1 |
| Qwen3.8-27B | email-Eu-core-main-n125-s0 | Direct construction | True | complete | 2 |
| Qwen3.8-27B | email-Eu-core-main-n150-s0 | Direct construction | True | complete | 2 |

## Appendix C. Complete Structural Query Results

All fourteen configurations are retained. Task-size macro gives each task and each of its five sizes equal weight. Task macro pools queries within a task before averaging the six tasks. Query-weighted accuracy gives degree half of the total weight. Invalid choices count as incorrect.

GPT-5.4 uses the no-reasoning configuration throughout the primary results: 733 correct and 800 legal answers. Its degree task has 364/400 correct, but equal averaging across sizes gives 91.4154%, rather than the pooled 91.0000%.

Table C.1. Aggregate accuracy (%) and response coverage.

| Configuration | Correct / 800 | Query weighted | Task macro | Task-size macro | Valid | Failed | U |
|---|---|---|---|---|---|---|---|
| Jev-1.13.0 | 609 | 76.1250 | 82.3750 | 82.5500 | 800 | 0 | 0 |
| Decider-2B | 289 | 36.1250 | 53.7083 | 53.7367 | 800 | 0 | 0 |
| Kev-4B | 344 | 43.0000 | 65.0000 | 65.0330 | 800 | 0 | 0 |
| Laya | 242 | 30.2500 | 43.5833 | 43.6056 | 800 | 0 | 0 |
| Q3.5-0.8B | 250 | 31.2500 | 44.0833 | 44.0984 | 800 | 0 | 0 |
| Q3.5-2B | 258 | 32.2500 | 44.0833 | 44.1249 | 800 | 0 | 0 |
| Q3.5-4B gen. | 406 | 50.7500 | 58.2500 | 58.3998 | 800 | 0 | 0 |
| Q3.5-9B | 406 | 50.7500 | 61.9167 | 62.0401 | 800 | 0 | 0 |
| Q3.5-27B | 590 | 73.7500 | 81.2500 | 81.3670 | 800 | 0 | 0 |
| Q3.8-27B | 533 | 66.6250 | 74.3750 | 74.4853 | 800 | 0 | 0 |
| Q2.5-72B | 608 | 76.0000 | 82.6667 | 82.8472 | 800 | 0 | 0 |
| Q3.5-4B score | 337 | 42.1250 | 58.7083 | 58.7545 | 800 | 0 | 0 |
| GPT-5.4 | <u>733</u> | <u>91.6250</u> | <u>92.0417</u> | <u>92.1109</u> | 800 | 0 | 0 |
| GPT-6-Astra | **800** | **100.0000** | **100.0000** | **100.0000** | 800 | 0 | 0 |

Ranked down each correctness/accuracy column; response-accounting columns are not ranked.

## C.2. All four sources

Table C.2. All four sources: task-size averaged accuracy (%).

| Configuration | Adjacency | Degree | Cycle | Pairwise connectivity | Distance | Articulation |
|---|---|---|---|---|---|---|
| Jev-1.13.0 | **100.0000** | 67.7999 | 87.5000 | 88.7500 | 90.0000 | 61.2500 |
| Decider-2B | 72.5000 | 9.9205 | 78.7500 | 50.0000 | 61.2500 | 50.0000 |
| Kev-4B | 88.7500 | 10.1982 | 85.0000 | 60.0000 | 83.7500 | 62.5000 |
| Laya | 50.0000 | 10.3838 | 50.0000 | 51.2500 | 50.0000 | 50.0000 |
| Q3.5-0.8B | 50.0000 | 12.0903 | 52.5000 | 50.0000 | 50.0000 | 50.0000 |
| Q3.5-2B | 50.0000 | 14.7494 | 50.0000 | 50.0000 | 50.0000 | 50.0000 |
| Q3.5-4B gen. | 78.7500 | 40.3990 | 80.0000 | 50.0000 | 50.0000 | 51.2500 |
| Q3.5-9B | 57.5000 | 34.7405 | 81.2500 | 73.7500 | 75.0000 | 50.0000 |
| Q3.5-27B | **100.0000** | 63.2020 | 87.5000 | 87.5000 | 80.0000 | 70.0000 |
| Q3.8-27B | 96.2500 | 55.6616 | 85.0000 | 86.2500 | 72.5000 | 51.2500 |
| Q2.5-72B | <u>98.7500</u> | 67.0833 | 78.7500 | 88.7500 | 91.2500 | 72.5000 |
| Q3.5-4B score | 90.0000 | 17.5271 | 78.7500 | 50.0000 | 61.2500 | 55.0000 |
| GPT-5.4 | **100.0000** | <u>91.4154</u> | <u>92.5000</u> | <u>96.2500</u> | <u>97.5000</u> | <u>75.0000</u> |
| GPT-6-Astra | **100.0000** | **100.0000** | **100.0000** | **100.0000** | **100.0000** | **100.0000** |

Ranked down each task column within the same source panel; higher accuracy is better.

## C.3. Facebook

Table C.3. Facebook: task-size averaged accuracy (%).

| Configuration | Adjacency | Degree | Cycle | Pairwise connectivity | Distance | Articulation |
|---|---|---|---|---|---|---|
| Jev-1.13.0 | **100.0000** | 62.9773 | **100.0000** | 90.0000 | 90.0000 | 60.0000 |
| Decider-2B | 70.0000 | 13.6869 | 95.0000 | 50.0000 | 60.0000 | 50.0000 |
| Kev-4B | 85.0000 | 14.7980 | **100.0000** | 50.0000 | 80.0000 | 65.0000 |
| Laya | 50.0000 | 14.5960 | 50.0000 | 55.0000 | 50.0000 | 50.0000 |
| Q3.5-0.8B | 50.0000 | 16.9015 | 65.0000 | 50.0000 | 50.0000 | 50.0000 |
| Q3.5-2B | 50.0000 | 15.2197 | 50.0000 | 50.0000 | 50.0000 | 50.0000 |
| Q3.5-4B gen. | 75.0000 | 34.7197 | 95.0000 | 50.0000 | 50.0000 | 50.0000 |
| Q3.5-9B | 55.0000 | 31.9066 | **100.0000** | 70.0000 | 85.0000 | 50.0000 |
| Q3.5-27B | **100.0000** | 67.2980 | 95.0000 | 95.0000 | 80.0000 | 65.0000 |
| Q3.8-27B | 95.0000 | 56.7854 | **100.0000** | 90.0000 | 75.0000 | 50.0000 |
| Q2.5-72B | **100.0000** | 66.5404 | 95.0000 | 95.0000 | 85.0000 | 60.0000 |
| Q3.5-4B score | 90.0000 | 19.4091 | 95.0000 | 50.0000 | 55.0000 | 55.0000 |
| GPT-5.4 | **100.0000** | 90.2727 | **100.0000** | **100.0000** | 95.0000 | 70.0000 |
| GPT-6-Astra | **100.0000** | **100.0000** | **100.0000** | **100.0000** | **100.0000** | **100.0000** |

Ranked down each task column within the same source panel; higher accuracy is better.

## C.4. ca-GrQc

Table C.4. ca-GrQc: task-size averaged accuracy (%).

| Configuration | Adjacency | Degree | Cycle | Pairwise connectivity | Distance | Articulation |
|---|---|---|---|---|---|---|
| Jev-1.13.0 | **100.0000** | 73.2045 | 95.0000 | 95.0000 | **100.0000** | 60.0000 |
| Decider-2B | 80.0000 | 12.0884 | 85.0000 | 50.0000 | 65.0000 | 50.0000 |
| Kev-4B | **100.0000** | 12.0884 | 90.0000 | 65.0000 | 75.0000 | 65.0000 |
| Laya | 50.0000 | 12.1641 | 50.0000 | 50.0000 | 50.0000 | 50.0000 |
| Q3.5-0.8B | 50.0000 | 13.2273 | 45.0000 | 50.0000 | 50.0000 | 50.0000 |
| Q3.5-2B | 50.0000 | 14.4495 | 50.0000 | 50.0000 | 50.0000 | 50.0000 |
| Q3.5-4B gen. | 75.0000 | 44.7146 | 85.0000 | 50.0000 | 50.0000 | 55.0000 |
| Q3.5-9B | 55.0000 | 44.3763 | 80.0000 | 80.0000 | 70.0000 | 50.0000 |
| Q3.5-27B | **100.0000** | 64.4141 | 85.0000 | 85.0000 | 90.0000 | 80.0000 |
| Q3.8-27B | **100.0000** | 52.5909 | 90.0000 | 85.0000 | 85.0000 | 50.0000 |
| Q2.5-72B | **100.0000** | 70.6288 | 85.0000 | **100.0000** | 95.0000 | 80.0000 |
| Q3.5-4B score | **100.0000** | 15.3864 | 85.0000 | 50.0000 | 65.0000 | 50.0000 |
| GPT-5.4 | **100.0000** | 87.1616 | **100.0000** | **100.0000** | 95.0000 | 85.0000 |
| GPT-6-Astra | **100.0000** | **100.0000** | **100.0000** | **100.0000** | **100.0000** | **100.0000** |

Ranked down each task column within the same source panel; higher accuracy is better.

## C.5. Power Grid

Table C.5. Power Grid: task-size averaged accuracy (%).

| Configuration | Adjacency | Degree | Cycle | Pairwise connectivity | Distance | Articulation |
|---|---|---|---|---|---|---|
| Jev-1.13.0 | **100.0000** | 63.8535 | 65.0000 | 80.0000 | 85.0000 | 65.0000 |
| Decider-2B | 80.0000 | 5.3535 | 55.0000 | 50.0000 | 55.0000 | 50.0000 |
| Kev-4B | 95.0000 | 5.3535 | 60.0000 | 60.0000 | 85.0000 | 65.0000 |
| Laya | 50.0000 | 7.2980 | 50.0000 | 50.0000 | 50.0000 | 50.0000 |
| Q3.5-0.8B | 50.0000 | 9.0960 | 50.0000 | 50.0000 | 50.0000 | 50.0000 |
| Q3.5-2B | 50.0000 | 15.3586 | 50.0000 | 50.0000 | 50.0000 | 50.0000 |
| Q3.5-4B gen. | 75.0000 | 42.0278 | 60.0000 | 50.0000 | 50.0000 | 50.0000 |
| Q3.5-9B | 55.0000 | 29.6364 | 65.0000 | 70.0000 | 60.0000 | 50.0000 |
| Q3.5-27B | **100.0000** | 56.8333 | 80.0000 | 90.0000 | 85.0000 | 65.0000 |
| Q3.8-27B | 95.0000 | 52.5631 | 65.0000 | 90.0000 | 70.0000 | 55.0000 |
| Q2.5-72B | **100.0000** | 65.5328 | 55.0000 | 75.0000 | 90.0000 | 75.0000 |
| Q3.5-4B score | 90.0000 | 18.0732 | 55.0000 | 50.0000 | 50.0000 | 60.0000 |
| GPT-5.4 | **100.0000** | 96.3485 | 85.0000 | 90.0000 | **100.0000** | 75.0000 |
| GPT-6-Astra | **100.0000** | **100.0000** | **100.0000** | **100.0000** | **100.0000** | **100.0000** |

Ranked down each task column within the same source panel; higher accuracy is better.

## C.6. Human PPI

Table C.6. Human PPI: task-size averaged accuracy (%).

| Configuration | Adjacency | Degree | Cycle | Pairwise connectivity | Distance | Articulation |
|---|---|---|---|---|---|---|
| Jev-1.13.0 | **100.0000** | 71.1641 | 90.0000 | 90.0000 | 85.0000 | 60.0000 |
| Decider-2B | 60.0000 | 8.5530 | 80.0000 | 50.0000 | 65.0000 | 50.0000 |
| Kev-4B | 75.0000 | 8.5530 | 90.0000 | 65.0000 | 95.0000 | 55.0000 |
| Laya | 50.0000 | 7.4773 | 50.0000 | 50.0000 | 50.0000 | 50.0000 |
| Q3.5-0.8B | 50.0000 | 9.1364 | 50.0000 | 50.0000 | 50.0000 | 50.0000 |
| Q3.5-2B | 50.0000 | 13.9697 | 50.0000 | 50.0000 | 50.0000 | 50.0000 |
| Q3.5-4B gen. | 90.0000 | 40.1338 | 80.0000 | 50.0000 | 50.0000 | 50.0000 |
| Q3.5-9B | 65.0000 | 33.0429 | 80.0000 | 75.0000 | 85.0000 | 50.0000 |
| Q3.5-27B | **100.0000** | 64.2626 | 90.0000 | 80.0000 | 65.0000 | 70.0000 |
| Q3.8-27B | 95.0000 | 60.7071 | 85.0000 | 80.0000 | 60.0000 | 50.0000 |
| Q2.5-72B | 95.0000 | 65.6313 | 80.0000 | 85.0000 | 95.0000 | 75.0000 |
| Q3.5-4B score | 80.0000 | 17.2399 | 80.0000 | 50.0000 | 75.0000 | 55.0000 |
| GPT-5.4 | **100.0000** | 91.8788 | 85.0000 | 95.0000 | **100.0000** | 70.0000 |
| GPT-6-Astra | **100.0000** | **100.0000** | **100.0000** | **100.0000** | **100.0000** | **100.0000** |

Ranked down each task column within the same source panel; higher accuracy is better.

### C.7. GPT-5.4 Source and Size Counts

Table C.7. GPT-5.4: correct/scheduled queries at each source and size.

| Source | n | Adjacency | Degree | Cycle | Pairwise connectivity | Distance | Articulation |
|---|---|---|---|---|---|---|---|
| Facebook | 8 | 4/4 | 14/16 | 4/4 | 4/4 | 4/4 | 4/4 |
| Facebook | 9 | 4/4 | 18/18 | 4/4 | 4/4 | 4/4 | 3/4 |
| Facebook | 10 | 4/4 | 18/20 | 4/4 | 4/4 | 4/4 | 2/4 |
| Facebook | 11 | 4/4 | 19/22 | 4/4 | 4/4 | 3/4 | 2/4 |
| Facebook | 12 | 4/4 | 21/24 | 4/4 | 4/4 | 4/4 | 3/4 |
| ca-GrQc | 8 | 4/4 | 14/16 | 4/4 | 4/4 | 4/4 | 2/4 |
| ca-GrQc | 9 | 4/4 | 17/18 | 4/4 | 4/4 | 4/4 | 4/4 |
| ca-GrQc | 10 | 4/4 | 16/20 | 4/4 | 4/4 | 4/4 | 3/4 |
| ca-GrQc | 11 | 4/4 | 19/22 | 4/4 | 4/4 | 3/4 | 4/4 |
| ca-GrQc | 12 | 4/4 | 21/24 | 4/4 | 4/4 | 4/4 | 4/4 |
| Power Grid | 8 | 4/4 | 16/16 | 4/4 | 4/4 | 4/4 | 4/4 |
| Power Grid | 9 | 4/4 | 18/18 | 4/4 | 4/4 | 4/4 | 4/4 |
| Power Grid | 10 | 4/4 | 19/20 | 3/4 | 3/4 | 4/4 | 3/4 |
| Power Grid | 11 | 4/4 | 20/22 | 3/4 | 4/4 | 4/4 | 2/4 |
| Power Grid | 12 | 4/4 | 23/24 | 3/4 | 3/4 | 4/4 | 2/4 |
| Human PPI | 8 | 4/4 | 16/16 | 4/4 | 4/4 | 4/4 | 3/4 |
| Human PPI | 9 | 4/4 | 18/18 | 3/4 | 4/4 | 4/4 | 3/4 |
| Human PPI | 10 | 4/4 | 19/20 | 4/4 | 4/4 | 4/4 | 3/4 |
| Human PPI | 11 | 4/4 | 16/22 | 3/4 | 4/4 | 4/4 | 2/4 |
| Human PPI | 12 | 4/4 | 22/24 | 3/4 | 3/4 | 4/4 | 3/4 |

## Appendix D. Complete Graph-Text Results

Each primary panel contains 100 targets or queries and five matched conditions. U denotes unsupported, not zero accuracy. ogbn-arxiv macro-F1 averages over the fixed forty classes. Paired intervals are pointwise 95% target-bootstrap intervals. Coverage cells give valid/failed/unsupported out of 100.

## D.1. ogbn-arxiv accuracy (%)

Table D.1. ogbn-arxiv accuracy (%).

| Configuration | T | G | TG | BAG | A |
|---|---|---|---|---|---|
| Jev-1.13.0 | 55.0000 | **61.0000** | 58.0000 | 57.0000 | 51.0000 |
| Decider-2B | 35.0000 | 45.0000 | 42.0000 | 38.0000 | 24.0000 |
| Kev-4B | 29.0000 | **61.0000** | 27.0000 | 27.0000 | 41.0000 |
| Laya | U | U | U | U | U |
| Q3.5-0.8B | 32.0000 | 34.0000 | 26.0000 | 24.0000 | 33.0000 |
| Q3.5-2B | 25.0000 | 12.0000 | 19.0000 | 23.0000 | 15.0000 |
| Q3.5-4B gen. | 21.0000 | 59.0000 | 30.0000 | 19.0000 | 42.0000 |
| Q3.5-9B | 32.0000 | 55.0000 | 41.0000 | 36.0000 | 39.0000 |
| Q3.5-27B | 46.0000 | **61.0000** | 58.0000 | 56.0000 | 49.0000 |
| Q3.8-27B | 62.0000 | 60.0000 | 71.0000 | 70.0000 | 51.0000 |
| Q2.5-72B | 51.0000 | 58.0000 | 59.0000 | 52.0000 | 45.0000 |
| Q3.5-4B score | U | U | U | U | U |
| GPT-5.4 | **73.0000** | **61.0000** | 73.0000 | 72.0000 | 48.0000 |
| GPT-6-Astra | **73.0000** | 60.0000 | **76.0000** | **73.0000** | **55.0000** |

Ranked down each condition column on this dataset's fixed targets; higher is better.

## D.2. ogbn-arxiv macro-F1 (%)

Table D.2. ogbn-arxiv macro-F1 (%).

| Configuration | T | G | TG | BAG | A |
|---|---|---|---|---|---|
| Jev-1.13.0 | 22.1632 | 25.8230 | 24.8080 | 24.5445 | 20.1305 |
| Decider-2B | 12.9338 | 17.6277 | 17.1978 | 13.3772 | 5.8756 |
| Kev-4B | 5.9031 | 25.4073 | 6.6961 | 6.5798 | 10.0644 |
| Laya | U | U | U | U | U |
| Q3.5-0.8B | 7.7786 | 8.0259 | 6.6967 | 7.2334 | 9.0932 |
| Q3.5-2B | 5.2578 | 3.7164 | 5.2396 | 4.4965 | 3.0843 |
| Q3.5-4B gen. | 6.5170 | 21.4803 | 11.0520 | 7.7386 | 14.1785 |
| Q3.5-9B | 13.0570 | 23.6818 | 17.3407 | 12.0195 | 13.5623 |
| Q3.5-27B | 18.6410 | 25.7326 | 23.7041 | 20.6193 | 18.5902 |
| Q3.8-27B | 22.2512 | **25.9381** | 29.7893 | 29.9108 | 19.4896 |
| Q2.5-72B | 20.9940 | 24.3895 | 25.5736 | 21.7584 | 19.5813 |
| Q3.5-4B score | U | U | U | U | U |
| GPT-5.4 | **35.8346** | 25.8859 | 30.3038 | 28.9699 | 13.8617 |
| GPT-6-Astra | 33.1572 | 24.9684 | **36.4209** | **35.4935** | **22.6972** |

Ranked down each condition column on this dataset's fixed targets; higher is better.

## D.3. ogbn-arxiv paired differences (pp), 95% intervals

Table D.3. ogbn-arxiv paired differences (pp), 95% intervals.

| Configuration | TG-BAG | G-A | TG-T | TG-G |
|---|---|---|---|---|
| Jev-1.13.0 | +1<br>[-4,+7] | +10<br>[+3,+18] | +3<br>[-5,+10] | -3<br>[-10,+4] |
| Decider-2B | +4<br>[+1,+8] | +21<br>[+12,+30] | +7<br>[+1,+13] | -3<br>[-14,+8] |
| Kev-4B | +0<br>[-3,+3] | +20<br>[+12,+29] | -2<br>[-10,+5] | -34<br>[-44,-24] |
| Laya | U | U | U | U |
| Q3.5-0.8B | +2<br>[-3,+8] | +1<br>[-5,+7] | -6<br>[-13,+0] | -8<br>[-14,-2] |
| Q3.5-2B | -4<br>[-10,+1] | -3<br>[-8,+1] | -6<br>[-15,+2] | +7<br>[+0,+14] |
| Q3.5-4B gen. | +11<br>[+6,+18] | +17<br>[+9,+25] | +9<br>[+1,+17] | -29<br>[-39,-20] |
| Q3.5-9B | +5<br>[-1,+11] | +16<br>[+9,+23] | +9<br>[+2,+17] | -14<br>[-23,-5] |
| Q3.5-27B | +2<br>[-3,+8] | +12<br>[+5,+20] | +12<br>[+4,+21] | -3<br>[-10,+5] |
| Q3.8-27B | +1<br>[-3,+6] | +9<br>[+3,+16] | +9<br>[+2,+16] | +11<br>[+5,+18] |
| Q2.5-72B | +7<br>[+2,+12] | +13<br>[+7,+20] | +8<br>[+3,+13] | +1<br>[-5,+7] |
| Q3.5-4B score | U | U | U | U |
| GPT-5.4 | +1<br>[-2, +4] | +13<br>[+3, +22] | +0<br>[-6, +5] | +12<br>[+4, +20] |
| GPT-6-Astra | +3<br>[+0,+7] | +5<br>[-2,+12] | +3<br>[-1,+7] | +16<br>[+9,+23] |

## D.4. ogbn-arxiv coverage

Table D.4. ogbn-arxiv coverage.

| Configuration | T | G | TG | BAG | A |
|---|---|---|---|---|---|
| Jev-1.13.0 | 100/0/0 | 100/0/0 | 100/0/0 | 100/0/0 | 100/0/0 |
| Decider-2B | 100/0/0 | 100/0/0 | 100/0/0 | 100/0/0 | 100/0/0 |
| Kev-4B | 100/0/0 | 100/0/0 | 100/0/0 | 100/0/0 | 100/0/0 |
| Laya | 0/0/100 | 0/0/100 | 0/0/100 | 0/0/100 | 0/0/100 |
| Q3.5-0.8B | 100/0/0 | 100/0/0 | 100/0/0 | 100/0/0 | 100/0/0 |
| Q3.5-2B | 100/0/0 | 100/0/0 | 100/0/0 | 100/0/0 | 100/0/0 |
| Q3.5-4B gen. | 100/0/0 | 100/0/0 | 100/0/0 | 100/0/0 | 100/0/0 |
| Q3.5-9B | 100/0/0 | 100/0/0 | 100/0/0 | 100/0/0 | 100/0/0 |
| Q3.5-27B | 100/0/0 | 100/0/0 | 100/0/0 | 100/0/0 | 100/0/0 |
| Q3.8-27B | 100/0/0 | 100/0/0 | 100/0/0 | 100/0/0 | 100/0/0 |
| Q2.5-72B | 100/0/0 | 100/0/0 | 100/0/0 | 100/0/0 | 100/0/0 |
| Q3.5-4B score | 0/0/100 | 0/0/100 | 0/0/100 | 0/0/100 | 0/0/100 |
| GPT-5.4 | 100/0/0 | 100/0/0 | 100/0/0 | 100/0/0 | 100/0/0 |
| GPT-6-Astra | 100/0/0 | 100/0/0 | 100/0/0 | 100/0/0 | 100/0/0 |

## D.5. STaRK-Prime Hit@1 (%)

Table D.5. STaRK-Prime Hit@1 (%).

| Configuration | T | G | TG | BAG | A* |
|---|---|---|---|---|---|
| Jev-1.13.0 | 48.0000 | 47.0000 | 49.0000 | 46.0000 | 42.0000 |
| Decider-2B | 20.0000 | 17.0000 | 18.0000 | 21.0000 | 20.0000 |
| Kev-4B | 39.0000 | 19.0000 | 26.0000 | 35.0000 | 34.0000 |
| Laya | U | U | U | U | U |
| Q3.5-0.8B | 9.0000 | 9.0000 | 9.0000 | 7.0000 | 8.0000 |
| Q3.5-2B | 16.0000 | 10.0000 | 16.0000 | 12.0000 | 13.0000 |
| Q3.5-4B gen. | 36.0000 | 17.0000 | 24.0000 | 28.0000 | 26.0000 |
| Q3.5-9B | 43.0000 | 25.0000 | 31.0000 | 38.0000 | 33.0000 |
| Q3.5-27B | 50.0000 | 44.0000 | 46.0000 | 43.0000 | 41.0000 |
| Q3.8-27B | 53.0000 | 50.0000 | 51.0000 | 51.0000 | 48.0000 |
| Q2.5-72B | 39.0000 | 35.0000 | 33.0000 | 35.0000 | 36.0000 |
| Q3.5-4B score | U | U | U | U | U |
| GPT-5.4 | 60.0000 | 51.0000 | 53.0000 | 56.0000 | 56.0000 |
| GPT-6-Astra | **69.0000** | **69.0000** | **67.0000** | **73.0000** | **71.0000** |

Ranked down each condition column on this dataset's fixed targets; higher is better.

## D.6. STaRK-Prime paired differences (pp), 95% intervals

Table D.6. STaRK-Prime paired differences (pp), 95% intervals.

| Configuration | TG-BAG | G-A* | TG-T | TG-G |
|---|---|---|---|---|
| Jev-1.13.0 | +3<br>[-6,+12] | +5<br>[-5,+14] | +1<br>[-8,+10] | +2<br>[-3,+8] |
| Decider-2B | -3<br>[-10,+4] | -3<br>[-13,+7] | -2<br>[-9,+4] | +1<br>[-5,+7] |
| Kev-4B | -9<br>[-17,-2] | -15<br>[-24,-6] | -13<br>[-22,-4] | +7<br>[+0,+15] |
| Laya | U | U | U | U |
| Q3.5-0.8B | +2<br>[-2,+6] | +1<br>[-4,+6] | +0<br>[-5,+5] | +0<br>[-4,+4] |
| Q3.5-2B | +4<br>[+0,+9] | -3<br>[-11,+4] | +0<br>[-5,+6] | +6<br>[+0,+13] |
| Q3.5-4B gen. | -4<br>[-11,+3] | -9<br>[-16,-1] | -12<br>[-21,-4] | +7<br>[+1,+13] |
| Q3.5-9B | -7<br>[-16,+0] | -8<br>[-18,+1] | -12<br>[-21,-4] | +6<br>[-2,+15] |
| Q3.5-27B | +3<br>[-6,+12] | +3<br>[-4,+10] | -4<br>[-12,+4] | +2<br>[-4,+8] |
| Q3.8-27B | +0<br>[-9,+8] | +2<br>[-7,+12] | -2<br>[-11,+7] | +1<br>[-4,+7] |
| Q2.5-72B | -2<br>[-8,+4] | -1<br>[-11,+9] | -6<br>[-14,+2] | -2<br>[-8,+5] |
| Q3.5-4B score | U | U | U | U |
| GPT-5.4 | -3<br>[-11, +4] | -5<br>[-13, +3] | -7<br>[-15, +1] | +2<br>[-4, +8] |
| GPT-6-Astra | -6<br>[-13,+1] | -2<br>[-9,+5] | -2<br>[-10,+5] | -2<br>[-6,+1] |

## D.7. STaRK-Prime coverage

Table D.7. STaRK-Prime coverage.

| Configuration | T | G | TG | BAG | A* |
|---|---|---|---|---|---|
| Jev-1.13.0 | 100/0/0 | 100/0/0 | 100/0/0 | 100/0/0 | 100/0/0 |
| Decider-2B | 100/0/0 | 100/0/0 | 100/0/0 | 100/0/0 | 100/0/0 |
| Kev-4B | 100/0/0 | 100/0/0 | 100/0/0 | 100/0/0 | 100/0/0 |
| Laya | 0/0/100 | 0/0/100 | 0/0/100 | 0/0/100 | 0/0/100 |
| Q3.5-0.8B | 100/0/0 | 100/0/0 | 100/0/0 | 100/0/0 | 100/0/0 |
| Q3.5-2B | 100/0/0 | 100/0/0 | 100/0/0 | 100/0/0 | 100/0/0 |
| Q3.5-4B gen. | 100/0/0 | 100/0/0 | 100/0/0 | 100/0/0 | 100/0/0 |
| Q3.5-9B | 100/0/0 | 100/0/0 | 100/0/0 | 100/0/0 | 100/0/0 |
| Q3.5-27B | 100/0/0 | 100/0/0 | 100/0/0 | 100/0/0 | 100/0/0 |
| Q3.8-27B | 100/0/0 | 100/0/0 | 100/0/0 | 100/0/0 | 100/0/0 |
| Q2.5-72B | 100/0/0 | 100/0/0 | 100/0/0 | 100/0/0 | 100/0/0 |
| Q3.5-4B score | 0/0/100 | 0/0/100 | 0/0/100 | 0/0/100 | 0/0/100 |
| GPT-5.4 | 100/0/0 | 100/0/0 | 98/2/0 | 100/0/0 | 100/0/0 |
| GPT-6-Astra | 100/0/0 | 100/0/0 | 100/0/0 | 100/0/0 | 100/0/0 |

## D.8. Independent 200-Target ogbn-arxiv Evaluation

This separate public-data evaluation is not pooled with either 100-target primary panel. It provides supporting accuracy results on its own 200-target set; it does not add targets to the primary paired comparisons.

Table D.8. Independent 200-target ogbn-arxiv accuracy (%).

| Configuration | T | G | TG | BAG | A |
|---|---|---|---|---|---|
| Decider-2B | 31.5000 | 44.0000 | 39.0000 | 35.0000 | 26.5000 |
| GPT-5.4 (reasoning reference) | <u>74.0000</u> | 58.5000 | <u>75.5000</u> | <u>74.0000</u> | <u>49.0000</u> |
| GPT-6-Astra | **74.5000** | **59.5000** | **77.5000** | **77.0000** | **51.0000** |
| Jev-1.13.0 | 54.5000 | <u>59.0000</u> | 57.5000 | 54.5000 | 47.5000 |
| Kev-4B | 28.5000 | 58.0000 | 28.0000 | 25.5000 | 37.0000 |
| Laya | U | U | U | U | U |
| Q3.5-27B | 48.0000 | 58.0000 | 56.0000 | 53.0000 | 45.5000 |
| Q3.5-4B gen. | 22.0000 | 56.0000 | 29.5000 | 21.0000 | 41.5000 |
| Q2.5-72B | 53.5000 | 56.5000 | 59.0000 | 56.5000 | 43.0000 |
| Q3.5-9B | 31.0000 | 54.0000 | 42.0000 | 38.5000 | 40.0000 |

Ranked down each condition column on this dataset's fixed targets; higher is better.

# Appendix E. Complete Sequential Results and Controls

Every configuration has ten scheduled graphs per task and construction mode. Direct construction supplies all legal next actions; heuristic-proposal selection supplies deduplicated named fixed-rule proposals. Means condition on feasible completion. Paired change is heuristic-proposal selection minus direct construction on jointly completed graphs; it is not a subtraction of marginal means when completed subsets differ. Matched random and classical controls use each mode's completed subset. Calls are those of the single evaluated trajectory. U and F mean unsupported and failed; -- indicates no defined mean.

## E.1. TSP

Table E.1. TSP: reference gaps (%), completion and paired change.

| Configuration | Direct construction Gap (done/10) | Heuristic-proposal selection Gap (done/10) | Paired change | Pairs |
|---|---|---|---|---|
| Jev-1.13.0 | 59.6641 (10/10) | 25.8731 (10/10) | -33.7911 | 10 |
| Decider-2B | 583.3950 (10/10) | 40.4541 (10/10) | -542.9409 | 10 |
| Kev-4B | 420.8638 (8/10) | 37.4815 (10/10) | -384.2287 | 8 |
| Laya | -- (0/10) | -- (0/10) | -- | 0 |
| Q3.5-0.8B | 576.7172 (10/10) | 33.8624 (10/10) | -542.8548 | 10 |
| Q3.5-2B | 539.5417 (10/10) | 34.7646 (10/10) | -504.7772 | 10 |
| Q3.5-4B gen. | 242.7966 (10/10) | 38.4328 (10/10) | -204.3638 | 10 |
| Q3.5-9B | 252.0041 (10/10) | 24.0653 (10/10) | -227.9388 | 10 |
| Q3.5-27B | 50.4023 (10/10) | 27.2948 (10/10) | -23.1075 | 10 |
| Q3.8-27B | 51.4061 (10/10) | 28.8632 (10/10) | -22.5429 | 10 |
| Q2.5-72B | 116.9987 (10/10) | 29.2820 (10/10) | -87.7167 | 10 |
| Q3.5-4B score | -- (0/10) | -- (0/10) | -- | 0 |
| GPT-5.4 | 29.4859 (10/10) | 32.2440 (10/10) | 2.7582 | 10 |
| GPT-6-Astra | **2.2117** (10/10) | **6.2354** (10/10) | 4.0237 | 10 |

Paired change: heuristic-proposal selection - direct construction; lower is better. Percent-task changes are percentage points; modularity changes use Q units. Ranked down each mode/task column only among full-ten-graph results; partial subsets remain unranked. Paired changes are diagnostics, not absolute-quality rankings.

Table E.2. TSP, Direct construction: objectives and matched controls.

| Configuration | Objective | Gap | Random gap | Classical gap | Done/U/F | Calls |
|---|---|---|---|---|---|---|
| Jev-1.13.0 | 43034.9000 | 59.6641 | 43.8850 | <u>5.7532</u> | 10/0/0 | 1257 |
| Decider-2B | 171889.2000 | 583.3950 | 43.8850 | <u>5.7532</u> | 10/0/0 | 1257 |
| Kev-4B | 140365.2500 | 420.8638 | <u>43.9812</u> | **5.0688** | 8/2/0 | 838 |
| Laya | -- | -- | -- | -- | 0/10/0 | 0 |
| Q3.5-0.8B | 172152.4000 | 576.7172 | 43.8850 | <u>5.7532</u> | 10/0/0 | 1257 |
| Q3.5-2B | 167230.3000 | 539.5417 | 43.8850 | <u>5.7532</u> | 10/0/0 | 1257 |
| Q3.5-4B gen. | 86291.9000 | 242.7966 | 43.8850 | <u>5.7532</u> | 10/0/0 | 1257 |
| Q3.5-9B | 91894.1000 | 252.0041 | 43.8850 | <u>5.7532</u> | 10/0/0 | 1257 |
| Q3.5-27B | 38203.7000 | 50.4023 | 43.8850 | <u>5.7532</u> | 10/0/0 | 1257 |
| Q3.8-27B | 38473.7000 | 51.4061 | 43.8850 | <u>5.7532</u> | 10/0/0 | 1257 |
| Q2.5-72B | 51646.5000 | 116.9987 | 43.8850 | <u>5.7532</u> | 10/0/0 | 1257 |
| Q3.5-4B score | -- | -- | -- | -- | 0/10/0 | 0 |
| GPT-5.4 | <u>34877.8000</u> | 29.4859 | 43.8850 | <u>5.7532</u> | 10/0/0 | 1257 |
| GPT-6-Astra | **27176.6000** | **2.2117** | 43.8850 | <u>5.7532</u> | 10/0/0 | 1257 |

Model calls include requests made before an incomplete or unsupported trajectory terminates; forced singleton actions require no call. Objectives: ranked down full-coverage model rows (TSP lower; other tasks higher). Gaps: ranked jointly across model/random/classical entries on the same ten graphs, lower is better. Each partial row instead ranks only its three matched gaps, never against other rows; partial objectives and calls are unranked.

Table E.3. TSP, Heuristic-proposal selection: objectives and matched controls.

| Configuration | Objective | Gap | Random gap | Classical gap | Done/U/F | Calls |
|---|---|---|---|---|---|---|
| Jev-1.13.0 | 32388.6000 | 25.8731 | 43.8850 | **5.7532** | 10/0/0 | 655 |
| Decider-2B | 38014.6000 | 40.4541 | 43.8850 | **5.7532** | 10/0/0 | 681 |
| Kev-4B | 37155.0000 | 37.4815 | 43.8850 | **5.7532** | 10/0/0 | 695 |
| Laya | -- | -- | -- | -- | 0/10/0 | 10 |
| Q3.5-0.8B | 34821.6000 | 33.8624 | 43.8850 | **5.7532** | 10/0/0 | 673 |
| Q3.5-2B | 36700.3000 | 34.7646 | 43.8850 | **5.7532** | 10/0/0 | 673 |
| Q3.5-4B gen. | 36498.6000 | 38.4328 | 43.8850 | **5.7532** | 10/0/0 | 686 |
| Q3.5-9B | <u>31982.7000</u> | 24.0653 | 43.8850 | **5.7532** | 10/0/0 | 658 |
| Q3.5-27B | 32620.6000 | 27.2948 | 43.8850 | **5.7532** | 10/0/0 | 655 |
| Q3.8-27B | 34598.1000 | 28.8632 | 43.8850 | **5.7532** | 10/0/0 | 680 |
| Q2.5-72B | 33904.7000 | 29.2820 | 43.8850 | **5.7532** | 10/0/0 | 647 |
| Q3.5-4B score | -- | -- | -- | -- | 0/10/0 | 10 |
| GPT-5.4 | 35026.2000 | 32.2440 | 43.8850 | **5.7532** | 10/0/0 | 668 |
| GPT-6-Astra | **28078.8000** | <u>6.2354</u> | 43.8850 | **5.7532** | 10/0/0 | 547 |

Model calls include requests made before an incomplete or unsupported trajectory terminates; forced singleton actions require no call. Objectives: ranked down full-coverage model rows (TSP lower; other tasks higher). Gaps: ranked jointly across model/random/classical entries on the same ten graphs, lower is better. Each partial row instead ranks only its three matched gaps, never against other rows; partial objectives and calls are unranked.

Table E.4. TSP: full-ten-graph control gaps (%).

| Control | Mean gap | Graphs |
|---|---|---|
| classical | **5.7532** | 10 |
| isolation_priority | 29.3081 | 10 |
| nearest | 25.8586 | 10 |
| one_step_lookahead | 52.2327 | 10 |
| random_candidate | 43.8850 | 10 |
| return_reserve | 15.5805 | 10 |

Control gaps are ranked on the same full ten-graph cohort; lower is better.

## E.2. Signed MaxCut

Table E.5. Signed MaxCut: reference gaps (%), completion and paired change.

| Configuration | Direct construction Gap (done/10) | Heuristic-proposal selection Gap (done/10) | Paired change | Pairs |
|---|---|---|---|---|
| Jev-1.13.0 | 99.9614 (10/10) | 9.0517 (10/10) | -90.9097 | 10 |
| Decider-2B | 99.8284 (10/10) | 19.0108 (10/10) | -80.8176 | 10 |
| Kev-4B | 99.8284 (10/10) | 21.9226 (10/10) | -77.9058 | 10 |
| Laya | -- (0/10) | -- (0/10) | -- | 0 |
| Q3.5-0.8B | 99.8054 (10/10) | 70.6073 (10/10) | -29.1981 | 10 |
| Q3.5-2B | 99.8284 (10/10) | 40.1330 (10/10) | -59.6954 | 10 |
| Q3.5-4B gen. | 99.8284 (10/10) | 12.3026 (10/10) | -87.5259 | 10 |
| Q3.5-9B | 99.3003 (10/10) | 19.9961 (10/10) | -79.3041 | 10 |
| Q3.5-27B | 104.0532 (10/10) | 10.4834 (10/10) | -93.5698 | 10 |
| Q3.8-27B | 99.8284 (10/10) | 23.1526 (10/10) | -76.6758 | 10 |
| Q2.5-72B | 105.9990 (10/10) | 11.5578 (10/10) | -94.4411 | 10 |
| Q3.5-4B score | -- (0/10) | -- (0/10) | -- | 0 |
| GPT-5.4 | 99.6673 (10/10) | 9.4116 (10/10) | -90.2557 | 10 |
| GPT-6-Astra | **6.3073** (10/10) | **7.7818** (10/10) | 1.4744 | 10 |

Paired change: heuristic-proposal selection - direct construction; lower is better. Percent-task changes are percentage points; modularity changes use Q units. Ranked down each mode/task column only among full-ten-graph results; partial subsets remain unranked. Paired changes are diagnostics, not absolute-quality rankings.

Table E.6. Signed MaxCut, Direct construction: objectives and matched controls.

| Configuration | Objective | Gap | Random gap | Classical gap | Done/U/F | Calls |
|---|---|---|---|---|---|---|
| Jev-1.13.0 | 0.0000 | 99.9614 | 82.7206 | 15.4096 | 10/0/0 | 1240 |
| Decider-2B | 0.2000 | 99.8284 | 82.7206 | 15.4096 | 10/0/0 | 1240 |
| Kev-4B | 0.2000 | 99.8284 | 82.7206 | 15.4096 | 10/0/0 | 1240 |
| Laya | -- | -- | -- | -- | 0/10/0 | 10 |
| Q3.5-0.8B | 0.2000 | 99.8054 | 82.7206 | 15.4096 | 10/0/0 | 1240 |
| Q3.5-2B | 0.2000 | 99.8284 | 82.7206 | 15.4096 | 10/0/0 | 1240 |
| Q3.5-4B gen. | 0.2000 | 99.8284 | 82.7206 | 15.4096 | 10/0/0 | 1240 |
| Q3.5-9B | 0.8000 | 99.3003 | 82.7206 | 15.4096 | 10/0/0 | 1240 |
| Q3.5-27B | -4.6000 | 104.0532 | 82.7206 | 15.4096 | 10/0/0 | 1240 |
| Q3.8-27B | 0.2000 | 99.8284 | 82.7206 | 15.4096 | 10/0/0 | 1240 |
| Q2.5-72B | -6.8000 | 105.9990 | 82.7206 | 15.4096 | 10/0/0 | 1240 |
| Q3.5-4B score | -- | -- | -- | -- | 0/10/0 | 10 |
| GPT-5.4 | 0.4000 | 99.6673 | 82.7206 | 15.4096 | 10/0/0 | 1240 |
| GPT-6-Astra | **103.6000** | **6.3073** | 82.7206 | 15.4096 | 10/0/0 | 1240 |

Model calls include requests made before an incomplete or unsupported trajectory terminates; forced singleton actions require no call. Objectives: ranked down full-coverage model rows (TSP lower; other tasks higher). Gaps: ranked jointly across model/random/classical entries on the same ten graphs, lower is better. Each partial row instead ranks only its three matched gaps, never against other rows; partial objectives and calls are unranked.

Table E.7. Signed MaxCut, Heuristic-proposal selection: objectives and matched controls.

| Configuration | Objective | Gap | Random gap | Classical gap | Done/U/F | Calls |
|---|---|---|---|---|---|---|
| Jev-1.13.0 | 100.6000 | 9.0517 | 82.7206 | 15.4096 | 10/0/0 | 955 |
| Decider-2B | 89.6000 | 19.0108 | 82.7206 | 15.4096 | 10/0/0 | 1016 |
| Kev-4B | 86.4000 | 21.9226 | 82.7206 | 15.4096 | 10/0/0 | 946 |
| Laya | -- | -- | -- | -- | 0/10/0 | 10 |
| Q3.5-0.8B | 32.6000 | 70.6073 | 82.7206 | 15.4096 | 10/0/0 | 1023 |
| Q3.5-2B | 66.4000 | 40.1330 | 82.7206 | 15.4096 | 10/0/0 | 1025 |
| Q3.5-4B gen. | 97.0000 | 12.3026 | 82.7206 | 15.4096 | 10/0/0 | 989 |
| Q3.5-9B | 88.6000 | 19.9961 | 82.7206 | 15.4096 | 10/0/0 | 1005 |
| Q3.5-27B | 99.0000 | 10.4834 | 82.7206 | 15.4096 | 10/0/0 | 982 |
| Q3.8-27B | 85.0000 | 23.1526 | 82.7206 | 15.4096 | 10/0/0 | 992 |
| Q2.5-72B | 97.8000 | 11.5578 | 82.7206 | 15.4096 | 10/0/0 | 1016 |
| Q3.5-4B score | -- | -- | -- | -- | 0/10/0 | 10 |
| GPT-5.4 | 100.2000 | 9.4116 | 82.7206 | 15.4096 | 10/0/0 | 987 |
| GPT-6-Astra | **102.0000** | **7.7818** | 82.7206 | 15.4096 | 10/0/0 | 945 |

Model calls include requests made before an incomplete or unsupported trajectory terminates; forced singleton actions require no call. Objectives: ranked down full-coverage model rows (TSP lower; other tasks higher). Gaps: ranked jointly across model/random/classical entries on the same ten graphs, lower is better. Each partial row instead ranks only its three matched gaps, never against other rows; partial objectives and calls are unranked.

Table E.8. Signed MaxCut: full-ten-graph control gaps (%).

| Control | Mean gap | Graphs |
| --- | --- | --- |
| assigned_gain | 19.7316 | 10 |
| classical | 15.4096 | 10 |
| degree_balance | 101.2925 | 10 |
| greedy_completion | **8.8727** | 10 |
| opposite_completion | 79.0745 | 10 |
| random_candidate | 82.7206 | 10 |

Control gaps are ranked on the same full ten-graph cohort; lower is better.

## E.3. Deterministic LT

Table E.9. Deterministic LT: reference gaps (%), completion and paired change.

| Configuration | Direct construction Gap (done/10) | Heuristic-proposal selection Gap (done/10) | Paired change | Pairs |
| --- | --- | --- | --- | --- |
| Jev-1.13.0 | 52.8359 (8/10) | 0.5000 (8/10) | -52.3359 | 8 |
| Decider-2B | 75.4013 (10/10) | 9.8681 (10/10) | -65.5332 | 10 |
| Kev-4B | 87.2393 (3/10) | 4.4118 (4/10) | -81.3570 | 3 |
| Laya | -- (0/10) | 0.0000 (1/10) | -- | 0 |
| Q3.5-0.8B | 84.1737 (10/10) | 16.3237 (10/10) | -67.8499 | 10 |
| Q3.5-2B | 77.6601 (10/10) | 22.4095 (10/10) | -55.2506 | 10 |
| Q3.5-4B gen. | 79.6852 (10/10) | 17.4011 (10/10) | -62.2841 | 10 |
| Q3.5-9B | 47.1873 (10/10) | 15.2482 (10/10) | -31.9391 | 10 |
| Q3.5-27B | 46.5092 (10/10) | 5.6717 (10/10) | -40.8375 | 10 |
| Q3.8-27B | 43.1821 (10/10) | 15.9568 (10/10) | -27.2253 | 10 |
| Q2.5-72B | 39.9037 (10/10) | 18.0717 (10/10) | -21.8320 | 10 |
| Q3.5-4B score | -- (0/10) | 0.0000 (1/10) | -- | 0 |
| GPT-5.4 | 33.4312 (10/10) | 5.4750 (10/10) | -27.9562 | 10 |
| GPT-6-Astra | **0.0000** (10/10) | **0.4000** (10/10) | 0.4000 | 10 |

Paired change: heuristic-proposal selection - direct construction; lower is better. Percent-task changes are percentage points; modularity changes use Q units. Ranked down each mode/task column only among full-ten-graph results; partial subsets remain unranked. Paired changes are diagnostics, not absolute-quality rankings.

Table E.10. Deterministic LT, Direct construction: objectives and matched controls.

| Configuration | Objective | Gap | Random gap | Classical gap | Done/U/F | Calls |
|---|---|---|---|---|---|---|
| Jev-1.13.0 | 11.3750 | 52.8359 | 21.9123 | **6.8438** | 8/0/2 | 18 |
| Decider-2B | 4.8000 | 75.4013 | 21.2475 | 5.4750 | 10/0/0 | 20 |
| Kev-4B | 3.0000 | 87.2393 | 13.9496 | **0.0000** | 3/7/0 | 13 |
| Laya | -- | -- | -- | -- | 0/10/0 | 0 |
| Q3.5-0.8B | 3.4000 | 84.1737 | 21.2475 | 5.4750 | 10/0/0 | 20 |
| Q3.5-2B | 4.4000 | 77.6601 | 21.2475 | 5.4750 | 10/0/0 | 20 |
| Q3.5-4B gen. | 4.4000 | 79.6852 | 21.2475 | 5.4750 | 10/0/0 | 20 |
| Q3.5-9B | 14.7000 | 47.1873 | 21.2475 | 5.4750 | 10/0/0 | 20 |
| Q3.5-27B | 14.6000 | 46.5092 | 21.2475 | 5.4750 | 10/0/0 | 20 |
| Q3.8-27B | 15.2000 | 43.1821 | 21.2475 | 5.4750 | 10/0/0 | 20 |
| Q2.5-72B | 15.6000 | 39.9037 | 21.2475 | 5.4750 | 10/0/0 | 20 |
| Q3.5-4B score | -- | -- | -- | -- | 0/10/0 | 0 |
| GPT-5.4 | 17.8000 | 33.4312 | 21.2475 | 5.4750 | 10/0/0 | 20 |
| GPT-6-Astra | **24.6000** | **0.0000** | 21.2475 | 5.4750 | 10/0/0 | 20 |

Model calls include requests made before an incomplete or unsupported trajectory terminates; forced singleton actions require no call. Objectives: ranked down full-coverage model rows (TSP lower; other tasks higher). Gaps: ranked jointly across model/random/classical entries on the same ten graphs, lower is better. Each partial row instead ranks only its three matched gaps, never against other rows; partial objectives and calls are unranked.

Table E.11. Deterministic LT, Heuristic-proposal selection: objectives and matched controls.

| Configuration | Objective | Gap | Random gap | Classical gap | Done/U/F | Calls |
|---|---|---|---|---|---|---|
| Jev-1.13.0 | 26.6250 | **0.5000** | 21.9123 | 6.8438 | 8/0/2 | 13 |
| Decider-2B | 23.0000 | 9.8681 | 21.2475 | 5.4750 | 10/0/0 | 13 |
| Kev-4B | 30.0000 | 4.4118 | 10.4622 | **0.0000** | 4/6/0 | 11 |
| Laya | 29.0000 | **0.0000** | **0.0000** | **0.0000** | 1/9/0 | 9 |
| Q3.5-0.8B | 21.3000 | 16.3237 | 21.2475 | 5.4750 | 10/0/0 | 12 |
| Q3.5-2B | 19.7000 | 22.4095 | 21.2475 | 5.4750 | 10/0/0 | 13 |
| Q3.5-4B gen. | 21.5000 | 17.4011 | 21.2475 | 5.4750 | 10/0/0 | 14 |
| Q3.5-9B | 21.4000 | 15.2482 | 21.2475 | 5.4750 | 10/0/0 | 14 |
| Q3.5-27B | 23.7000 | 5.6717 | 21.2475 | 5.4750 | 10/0/0 | 14 |
| Q3.8-27B | 21.8000 | 15.9568 | 21.2475 | 5.4750 | 10/0/0 | 13 |
| Q2.5-72B | 20.6000 | 18.0717 | 21.2475 | 5.4750 | 10/0/0 | 14 |
| Q3.5-4B score | 29.0000 | **0.0000** | **0.0000** | **0.0000** | 1/9/0 | 9 |
| GPT-5.4 | 23.4000 | 5.4750 | 21.2475 | 5.4750 | 10/0/0 | 12 |
| GPT-6-Astra | **24.5000** | **0.4000** | 21.2475 | 5.4750 | 10/0/0 | 13 |

Model calls include requests made before an incomplete or unsupported trajectory terminates; forced singleton actions require no call. Objectives: ranked down full-coverage model rows (TSP lower; other tasks higher). Gaps: ranked jointly across model/random/classical entries on the same ten graphs, lower is better. Each partial row instead ranks only its three matched gaps, never against other rows; partial objectives and calls are unranked.

Table E.12. Deterministic LT: full-ten-graph control gaps (%).

| Control | Mean gap | Graphs |
|---|---|---|
| classical | **5.4750** | 10 |
| marginal_cascade | **5.4750** | 10 |
| outdegree | 30.6933 | 10 |
| random_candidate | 21.2475 | 10 |
| threshold_deficit | 20.2482 | 10 |
| two_hop_reach | 31.7408 | 10 |

Control gaps are ranked on the same full ten-graph cohort; lower is better.

## E.4. Binary modularity

Table E.13. Binary modularity: reference gaps (Q), completion and paired change.

| Configuration | Direct construction Gap (done/10) | Heuristic-proposal selection Gap (done/10) | Paired change | Pairs |
|---|---|---|---|---|
| Jev-1.13.0 | 0.4783832 (10/10) | 0.0555132 (10/10) | -0.4228700 | 10 |
| Decider-2B | 0.4783832 (10/10) | 0.1465732 (10/10) | -0.3318100 | 10 |
| Kev-4B | 0.4783832 (10/10) | 0.0964632 (10/10) | -0.3819200 | 10 |
| Laya | 0.5105148 (3/10) | -- (0/10) | -- | 0 |
| Q3.5-0.8B | 0.4783832 (10/10) | 0.1936732 (10/10) | -0.2847100 | 10 |
| Q3.5-2B | 0.4783832 (10/10) | 0.3656432 (10/10) | -0.1127400 | 10 |
| Q3.5-4B gen. | 0.4654832 (10/10) | 0.0885532 (10/10) | -0.3769300 | 10 |
| Q3.5-9B | 0.4783832 (10/10) | 0.1158132 (10/10) | -0.3625700 | 10 |
| Q3.5-27B | 0.4783832 (10/10) | 0.0892932 (10/10) | -0.3890900 | 10 |
| Q3.8-27B | 0.4752832 (10/10) | 0.1086932 (10/10) | -0.3665900 | 10 |
| Q2.5-72B | 0.3758732 (10/10) | 0.0849032 (10/10) | -0.2909700 | 10 |
| Q3.5-4B score | 0.4804942 (4/10) | 0.1562265 (1/10) | -0.3255000 | 1 |
| GPT-5.4 | 0.4042398 (10/10) | 0.0422862 (10/10) | -0.3619537 | 10 |
| GPT-6-Astra | **0.0002611** (10/10) | **0.0129828** (10/10) | 0.0127217 | 10 |

Paired change: heuristic-proposal selection - direct construction; lower is better. Percent-task changes are percentage points; modularity changes use Q units. Ranked down each mode/task column only among full-ten-graph results; partial subsets remain unranked. Paired changes are diagnostics, not absolute-quality rankings.

Table E.14. Binary modularity, Direct construction: objectives and matched controls.

| Configuration | Objective | Gap | Random gap | Classical gap | Done/U/F | Calls |
|---|---|---|---|---|---|---|
| Jev-1.13.0 | 0.0000000 | 0.4783832 | 0.4175 | 0.4784 | 10/0/0 | 990 |
| Decider-2B | 0.0000000 | 0.4783832 | 0.4175 | 0.4784 | 10/0/0 | 990 |
| Kev-4B | 0.0000000 | 0.4783832 | 0.4175 | 0.4784 | 10/0/0 | 990 |
| Laya | -0.0266000 | 0.5105148 | **0.4102** | 0.4839 | 3/7/0 | 186 |
| Q3.5-0.8B | 0.0000000 | 0.4783832 | 0.4175 | 0.4784 | 10/0/0 | 990 |
| Q3.5-2B | 0.0000000 | 0.4783832 | 0.4175 | 0.4784 | 10/0/0 | 990 |
| Q3.5-4B gen. | 0.0129000 | 0.4654832 | 0.4175 | 0.4784 | 10/0/0 | 990 |
| Q3.5-9B | 0.0000000 | 0.4783832 | 0.4175 | 0.4784 | 10/0/0 | 990 |
| Q3.5-27B | 0.0000000 | 0.4783832 | 0.4175 | 0.4784 | 10/0/0 | 990 |
| Q3.8-27B | 0.0031000 | 0.4752832 | 0.4175 | 0.4784 | 10/0/0 | 990 |
| Q2.5-72B | 0.1025000 | 0.3758732 | 0.4175 | 0.4784 | 10/0/0 | 990 |
| Q3.5-4B score | 0.0000000 | 0.4804942 | **0.4111** | 0.4805 | 4/6/0 | 265 |
| GPT-5.4 | 0.0741434 | 0.4042398 | 0.4175 | 0.4784 | 10/0/0 | 990 |
| GPT-6-Astra | **0.4781000** | **0.0002611** | 0.4175 | 0.4784 | 10/0/0 | 990 |

Model calls include requests made before an incomplete or unsupported trajectory terminates; forced singleton actions require no call. Objectives: ranked down full-coverage model rows (TSP lower; other tasks higher). Gaps: ranked jointly across model/random/classical entries on the same ten graphs, lower is better. Each partial row instead ranks only its three matched gaps, never against other rows; partial objectives and calls are unranked.

Table E.15. Binary modularity, Heuristic-proposal selection: objectives and matched controls.

| Configuration | Objective | Gap | Random gap | Classical gap | Done/U/F | Calls |
|---|---|---|---|---|---|---|
| Jev-1.13.0 | 0.4229000 | 0.0555132 | 0.4175 | 0.4784 | 10/0/0 | 737 |
| Decider-2B | 0.3318000 | 0.1465732 | 0.4175 | 0.4784 | 10/0/0 | 729 |
| Kev-4B | 0.3819000 | 0.0964632 | 0.4175 | 0.4784 | 10/0/0 | 729 |
| Laya | -- | -- | -- | -- | 0/10/0 | 10 |
| Q3.5-0.8B | 0.2847000 | 0.1936732 | 0.4175 | 0.4784 | 10/0/0 | 778 |
| Q3.5-2B | 0.1127000 | 0.3656432 | 0.4175 | 0.4784 | 10/0/0 | 750 |
| Q3.5-4B gen. | 0.3898000 | 0.0885532 | 0.4175 | 0.4784 | 10/0/0 | 727 |
| Q3.5-9B | 0.3626000 | 0.1158132 | 0.4175 | 0.4784 | 10/0/0 | 769 |
| Q3.5-27B | 0.3891000 | 0.0892932 | 0.4175 | 0.4784 | 10/0/0 | 706 |
| Q3.8-27B | 0.3697000 | 0.1086932 | 0.4175 | 0.4784 | 10/0/0 | 745 |
| Q2.5-72B | 0.3935000 | 0.0849032 | 0.4175 | 0.4784 | 10/0/0 | 725 |
| Q3.5-4B score | 0.3255000 | **0.1562265** | 0.4316 | 0.4817 | 1/9/0 | 48 |
| GPT-5.4 | 0.4360970 | 0.0422862 | 0.4175 | 0.4784 | 10/0/0 | 704 |
| GPT-6-Astra | **0.4654000** | **0.0129828** | 0.4175 | 0.4784 | 10/0/0 | 743 |

Model calls include requests made before an incomplete or unsupported trajectory terminates; forced singleton actions require no call. Objectives: ranked down full-coverage model rows (TSP lower; other tasks higher). Gaps: ranked jointly across model/random/classical entries on the same ten graphs, lower is better. Each partial row instead ranks only its three matched gaps, never against other rows; partial objectives and calls are unranked.

Table E.16. Binary modularity: full-ten-graph control gaps (Q).

| Control | Mean gap | Graphs |
|---|---|---|
| assigned_gain | 0.1202 | 10 |
| classical | 0.4784 | 10 |
| degree_balance | 0.5080 | 10 |
| greedy_completion | **0.0361** | 10 |
| opposite_completion | 0.4064 | 10 |
| random_candidate | 0.4175 | 10 |

Control gaps are ranked on the same full ten-graph cohort; lower is better.

Modularity uses the rational Q* references in Appendix G.1. GPT-5.4's per-graph full-precision objectives and Astra's supplied high-precision aggregate gaps avoid the four-decimal input rounding of the other configurations. For the latter, objective rounding contributes at most 0.00005 to a mode mean and 0.0001 to a paired change. Tiny signed per-graph residuals within that rounding range do not establish an improvement over Q*.

## E.5. Cross-Task Quality and Completion

Table E.17. Direct construction: reference gaps and completion.

| Configuration | TSP gap (%) | MaxCut gap (%) | LT gap (%) | Modularity gap (Q) |
|---|---|---|---|---|
| Jev-1.13.0 | 59.6641 (10/10) | 99.9614 (10/10) | 52.8359 (8/10) | 0.4783832 (10/10) |
| Decider-2B | 583.3950 (10/10) | 99.8284 (10/10) | 75.4013 (10/10) | 0.4783832 (10/10) |
| Kev-4B | 420.8638 (8/10) | 99.8284 (10/10) | 87.2393 (3/10) | 0.4783832 (10/10) |
| Laya | -- (0/10) | -- (0/10) | -- (0/10) | 0.5105148 (3/10) |
| Q3.5-0.8B | 576.7172 (10/10) | 99.8054 (10/10) | 84.1737 (10/10) | 0.4783832 (10/10) |
| Q3.5-2B | 539.5417 (10/10) | 99.8284 (10/10) | 77.6601 (10/10) | 0.4783832 (10/10) |
| Q3.5-4B gen. | 242.7966 (10/10) | 99.8284 (10/10) | 79.6852 (10/10) | 0.4654832 (10/10) |
| Q3.5-9B | 252.0041 (10/10) | 99.3003 (10/10) | 47.1873 (10/10) | 0.4783832 (10/10) |
| Q3.5-27B | 50.4023 (10/10) | 104.0532 (10/10) | 46.5092 (10/10) | 0.4783832 (10/10) |
| Q3.8-27B | 51.4061 (10/10) | 99.8284 (10/10) | 43.1821 (10/10) | 0.4752832 (10/10) |
| Q2.5-72B | 116.9987 (10/10) | 105.9990 (10/10) | 39.9037 (10/10) | 0.3758732 (10/10) |
| Q3.5-4B score | -- (0/10) | -- (0/10) | -- (0/10) | 0.4804942 (4/10) |
| GPT-5.4 | 29.4859 (10/10) | 99.6673 (10/10) | 33.4312 (10/10) | 0.4042398 (10/10) |
| GPT-6-Astra | **2.2117** (10/10) | **6.3073** (10/10) | **0.0000** (10/10) | **0.0002611** (10/10) |

Cells show gap (completed/scheduled); all gaps are lower-is-better. -- denotes no feasible completion, not zero. Ranked down each mode/task column only among full-ten-graph results; partial subsets remain unranked.

Table E.18. Heuristic-proposal selection: reference gaps and completion.

| Configuration | TSP gap (%) | MaxCut gap (%) | LT gap (%) | Modularity gap (Q) |
|---|---|---|---|---|
| Jev-1.13.0 | 25.8731 (10/10) | 9.0517 (10/10) | 0.5000 (8/10) | 0.0555132 (10/10) |
| Decider-2B | 40.4541 (10/10) | 19.0108 (10/10) | 9.8681 (10/10) | 0.1465732 (10/10) |
| Kev-4B | 37.4815 (10/10) | 21.9226 (10/10) | 4.4118 (4/10) | 0.0964632 (10/10) |
| Laya | -- (0/10) | -- (0/10) | 0.0000 (1/10) | -- (0/10) |
| Q3.5-0.8B | 33.8624 (10/10) | 70.6073 (10/10) | 16.3237 (10/10) | 0.1936732 (10/10) |
| Q3.5-2B | 34.7646 (10/10) | 40.1330 (10/10) | 22.4095 (10/10) | 0.3656432 (10/10) |
| Q3.5-4B gen. | 38.4328 (10/10) | 12.3026 (10/10) | 17.4011 (10/10) | 0.0885532 (10/10) |
| Q3.5-9B | 24.0653 (10/10) | 19.9961 (10/10) | 15.2482 (10/10) | 0.1158132 (10/10) |
| Q3.5-27B | 27.2948 (10/10) | 10.4834 (10/10) | 5.6717 (10/10) | 0.0892932 (10/10) |
| Q3.8-27B | 28.8632 (10/10) | 23.1526 (10/10) | 15.9568 (10/10) | 0.1086932 (10/10) |
| Q2.5-72B | 29.2820 (10/10) | 11.5578 (10/10) | 18.0717 (10/10) | 0.0849032 (10/10) |
| Q3.5-4B score | -- (0/10) | -- (0/10) | 0.0000 (1/10) | 0.1562265 (1/10) |
| GPT-5.4 | 32.2440 (10/10) | 9.4116 (10/10) | 5.4750 (10/10) | 0.0422862 (10/10) |
| GPT-6-Astra | **6.2354** (10/10) | **7.7818** (10/10) | **0.4000** (10/10) | **0.0129828** (10/10) |

Cells show gap (completed/scheduled); all gaps are lower-is-better. -- denotes no feasible completion, not zero. Ranked down each mode/task column only among full-ten-graph results; partial subsets remain unranked.

# Appendix F. Interface Details and Supplementary Graph-Text Evaluation

## F.1 Interface Readouts and Candidate Scoring

The implementation specification distinguishes native selection from generation and candidate scoring. Decider uses a restricted-letter one-pass BF16 readout with temperature 1.164. Kev uses an FP32 pointer head on Qwen3.5-4B-Base with merged LoRA and temperature approximately 2.40605. Laya uses an FP32 encoder/choice head with at most twelve options and 1,024 tokens. Jev's serving precision is undisclosed. The evaluated Qwen candidate-scoring implementation runs on GPU and uses the conditional argmax over supplied candidate-token probabilities; its one generated token is counted but not used for the choice. At a matched task state, generation and scoring use the same task prompt and candidate options but differ in output readout; sequential trajectories may reach different states.

## F.2 200-Target ogbn-arxiv Evaluation Protocol

This section specifies the protocol, input lengths and uncertainty estimates for the separate 200-target ogbn-arxiv cohort in Table A.5. Its model-interface configurations are those listed in Table F.2.

The ogbn-arxiv [8] snapshot contains 169,343 papers and 1,166,243 directed citation edges. Official temporal splits contain 90,941 training papers (through 2017), 29,799 validation papers (2018) and 48,603 test papers (2019 onward). Context is drawn from the full static graph under transductive evaluation. Seed 20261001 selects 200 targets without replacement from sorted official test IDs, independently of labels. Seed 20261002 selects 12 validation targets for interface-support checks, without performance-based prompt or model tuning.

Titles and abstracts are joined to graph nodes by MAG paper ID. The text collection's 179,719 unique records cover all 1,612 selected distinct nodes. Unicode whitespace is collapsed, and each node is capped at 256 title characters and 1,600 abstract characters. All models receive the same resulting strings.

For each target, select up to four training neighbors from the union of incoming and outgoing neighbors and exactly four training non-neighbors. Selection and presentation order use deterministic SHA-256 rankings with seed 20261001 and separate neighbor, control and display domains; ties use numeric node IDs. Neighbor shortages remain, yielding four to eight context nodes. All induced directed edges are retained. Context identifiers are anonymized, and original IDs and neighbor/control roles are hidden. Training-node class labels are shared across all conditions; held-out target labels are excluded.

All five conditions share training-label anchors. T adds target text, G edges, TG target/context text and edges, BAG target/context text without edges, and A anonymous anchors and labels only. The question asks for exactly one of forty class IDs; a directed pair [u,v] means u cites v. Omitted text or edges indicate unavailable information, not evidence of absence.

Table F.1. Fixed class descriptions; inference follows numeric ID order.

| ID | Description | ID | Description |
|---|---|---|---|
| 0 | arxiv cs na | 10 | arxiv cs ai |
| 1 | arxiv cs mm | 11 | arxiv cs ma |
| 2 | arxiv cs lo | 12 | arxiv cs gl |
| 3 | arxiv cs cy | 13 | arxiv cs ne |
| 4 | arxiv cs cr | 14 | arxiv cs sc |
| 5 | arxiv cs dc | 15 | arxiv cs ar |
| 6 | arxiv cs hc | 16 | arxiv cs cv |
| 7 | arxiv cs ce | 17 | arxiv cs gr |
| 8 | arxiv cs ni | 18 | arxiv cs et |
| 9 | arxiv cs cc | 19 | arxiv cs sy |
| 20 | arxiv cs cg | 30 | arxiv cs cl |
| 21 | arxiv cs oh | 31 | arxiv cs ir |
| 22 | arxiv cs pl | 32 | arxiv cs ms |
| 23 | arxiv cs se | 33 | arxiv cs fl |
| 24 | arxiv cs lg | 34 | arxiv cs ds |
| 25 | arxiv cs sd | 35 | arxiv cs os |
| 26 | arxiv cs si | 36 | arxiv cs gt |
| 27 | arxiv cs ro | 37 | arxiv cs db |
| 28 | arxiv cs it | 38 | arxiv cs dl |
| 29 | arxiv cs pf | 39 | arxiv cs dm |

Table F.2. Independent 200-target token limits and maximum observed input lengths.

| Configuration | Maximum observed input | Context limit | Configured output value | Reasoning mode / metadata |
|---|---|---|---|---|
| Jev | 4564 | Undisclosed | Native choice | N/A |
| Decider-2B | 3987 | 32768 | Native choice | N/A |
| Kev-4B | 3897 | 8192 | Native choice | N/A |
| Qwen3.5-4B | 4149 | 16384 | 16 | Thinking disabled |
| Qwen3.5-9B | 4149 | 16384 | 16 | Thinking disabled |
| Qwen3.5-27B | 4149 | 16384 | 16 | Thinking disabled |
| Qwen2.5-72B-Instruct | 4154 | 16384 | 16 | Standard instruct |
| GPT-5.4 (reasoning reference) | 4104 | Undisclosed | 4096 | Medium |
| GPT-6-Astra | 4104 | Undisclosed | 4096 | Medium |

For the GPT rows in Table F.2, Medium denotes observed response metadata rather than an explicitly submitted reasoning setting. The output value of 4,096 is a configured value, not a verified limit on total reasoning tokens.

The nine supported configurations complete 60 validation decisions and 1,000 test decisions each: 540 validation and 9,000 test decisions overall. Laya is unsupported. Jev's 18 validation and 287 test probability vectors with nonunit sums retain their otherwise legal selected answers. Validation is used for interface checks, not included in test accuracy.

### F.3 200-Target ogbn-arxiv Macro-F1 and Paired Intervals

Macro-F1 uses all 40 classes and is multiplied by 100; classes absent from both truth and prediction contribute zero. Accuracy changes are in percentage points, while macro-F1 changes are multiplied by 100. Both analyses use the same 200 target pairs and unadjusted 95% paired target-bootstrap intervals. Dashes denote unavailable results.

The original paired analysis uses 1,000 target resamples, seed 20261001, and sorted zero-based endpoints 24 and 974 for the 95% interval. Repeated targets keep their paired conditions together. The 1,408 test-anchor slots contain 1,321 distinct training anchors, so these target-level intervals do not capture all network dependence.

Table F.3. Independent 200-target fixed-forty-class macro-F1 (%).

| Configuration | T | G | TG | BAG | A |
|---|---|---|---|---|---|
| Jev | 27.8 | <u>32.9</u> | 32.5 | 29.5 | <u>20.0</u> |
| Decider-2B | 12.7 | 27.1 | 18.4 | 15.6 | 9.8 |
| Kev-4B | 7.6 | 30.2 | 11.2 | 10.8 | 11.2 |
| Laya (unsupported) | — | — | — | — | — |
| Qwen3.5-4B | 9.4 | 25.7 | 16.0 | 13.7 | 15.2 |
| Qwen3.5-9B | 15.5 | 27.2 | 20.2 | 15.4 | 15.7 |
| Qwen3.5-27B | 24.8 | 32.5 | 30.3 | 27.3 | 19.2 |
| Qwen2.5-72B-Instruct | 28.6 | 31.1 | 32.3 | 31.6 | 19.8 |
| GPT-5.4 (reasoning reference) | **46.5** | **33.2** | <u>46.6</u> | <u>46.2</u> | 19.6 |
| GPT-6-Astra | <u>44.0</u> | 31.9 | **47.4** | **46.7** | **22.7** |

Ranked down each condition column on this dataset's fixed targets; higher is better.

Table F.4. Independent 200-target paired accuracy changes (pp), with 95% intervals.

| Configuration | TG-T | TG-G | TG-BAG | G-A |
|---|---|---|---|---|
| Jev | +3.0 [-1.5,+7.5] | -1.5 [-6.0,+3.0] | +3.0 [-1.0,+7.0] | +11.5 [+6.0,+17.0] |
| Decider-2B | +7.5 [+2.5,+12.5] | -5.0 [-12.0,+1.5] | +4.0 [+1.5,+7.0] | +17.5 [+11.0,+23.5] |
| Kev-4B | -0.5 [-5.0,+4.0] | -30.0 [-37.0,-23.0] | +2.5 [+0.0,+5.5] | +21.0 [+15.5,+27.5] |
| Qwen3.5-4B | +7.5 [+1.5,+13.5] | -26.5 [-33.5,-19.5] | +8.5 [+4.5,+13.0] | +14.5 [+9.0,+20.5] |
| Qwen3.5-9B | +11.0 [+5.5,+16.0] | -12.0 [-18.5,-6.0] | +3.5 [+0.0,+7.0] | +14.0 [+8.5,+19.0] |
| Qwen3.5-27B | +8.0 [+1.5,+14.0] | -2.0 [-7.5,+3.5] | +3.0 [-0.5,+6.5] | +12.5 [+7.5,+18.0] |
| Qwen2.5-72B-Instruct | +5.5 [+2.0,+9.5] | +2.5 [-1.5,+6.5] | +2.5 [-0.5,+6.0] | +13.5 [+9.0,+18.5] |
| GPT-5.4 (reasoning reference) | +1.5 [-3.0,+5.5] | +17.0 [+12.0,+23.0] | +1.5 [-1.5,+4.5] | +9.5 [+3.5,+16.0] |
| GPT-6-Astra | +3.0 [-1.0,+7.0] | +18.0 [+12.5,+23.5] | +0.5 [-2.0,+3.0] | +8.5 [+3.0,+14.0] |

Table F.5. Independent 200-target paired macro-F1 changes (×100), with 95% intervals.

| Configuration | TG-T | TG-G | TG-BAG | G-A |
|---|---|---|---|---|
| Jev | +4.6 [-1.2,+6.6] | -0.4 [-3.9,+2.7] | +2.9 [-0.7,+4.8] | +12.9 [+5.8,+15.7] |
| Decider-2B | +5.8 [+0.7,+9.4] | -8.7 [-12.4,-1.5] | +2.8 [+0.2,+4.8] | +17.3 [+7.5,+20.2] |
| Kev-4B | +3.6 [-0.4,+5.9] | -18.9 [-22.3,-10.3] | +0.5 [+0.1,+1.0] | +19.0 [+10.6,+21.9] |
| Qwen3.5-4B | +6.6 [+1.8,+10.4] | -9.7 [-14.1,-5.1] | +2.3 [+0.0,+4.8] | +10.5 [+5.3,+14.0] |
| Qwen3.5-9B | +4.7 [-1.1,+10.5] | -7.0 [-11.6,-1.1] | +4.9 [+0.9,+7.3] | +11.5 [+4.9,+14.8] |
| Qwen3.5-27B | +5.5 [-0.3,+9.0] | -2.2 [-5.9,+4.1] | +3.0 [+0.2,+5.0] | +13.3 [+6.4,+16.2] |
| Qwen2.5-72B-Instruct | +3.7 [-0.0,+7.5] | +1.2 [-0.6,+3.3] | +0.7 [-2.8,+3.9] | +11.3 [+5.1,+13.7] |
| GPT-5.4 (reasoning reference) | +0.1 [-2.8,+3.2] | +13.4 [+7.3,+16.7] | +0.3 [-2.3,+2.7] | +13.6 [+5.8,+16.4] |
| GPT-6-Astra | +3.5 [-0.4,+5.7] | +15.5 [+8.7,+19.9] | +0.7 [-1.4,+1.8] | +9.2 [+3.1,+11.7] |

Jev's TG-BAG accuracy difference is +3.0 percentage points [-1.0, +7.0], leaving its explicit-edge benefit unresolved by this interval. Decider and Qwen3.5-4B have positive TG-BAG accuracy intervals; the other seven configurations, including Jev, touch or cross zero. Every G-A accuracy interval is positive, with point estimates of 8.5–21.0 percentage points. The positive G-A contrasts show the value of edges when only label anchors are otherwise available. GPT-5.4 (reasoning reference) and Astra achieve TG accuracies of 75.5% and 77.5%, respectively.

Table F.6. Independent 1,000-test-call token usage.

| Configuration | Input tokens | Output tokens | Reasoning tokens |
|---|---|---|---|
| Jev | 2,093,713 | 327,869 | — |
| Decider-2B | 1,669,548 | 0 | — |
| Kev-4B | 1,579,548 | 0 | — |
| Qwen3.5-4B | 1,777,549 | 2,817 | 0 |
| Qwen3.5-9B | 1,777,549 | 2,878 | 0 |
| Qwen3.5-27B | 1,777,549 | 2,860 | 0 |
| Qwen2.5-72B-Instruct | 1,774,761 | 2,907 | — |
| GPT-5.4 (reasoning reference) | 1,753,284 | 205,790 | 198,790 |
| GPT-6-Astra | 1,753,284 | 30,532 | 24,600 |

## Appendix G. Optimization Reference Calculations

The references below apply to the fixed benchmark instances, not the full source networks or other partition and diffusion variants. TSP retains published proven optima and MaxCut retains historical best-known values; this appendix specifies the two induced-graph tasks.

### G.1 Binary Modularity

The ten induced instances from SNAP ca-GrQc [35] are simple, undirected and unweighted, with resolution $\gamma$=1 and at most two communities, without size or volume balance constraints. Let m be the number of edges, $d_v$ the degree of vertex v, $x_v$ indicate membership in A, $V_A=\Sigma_v d_v x_v$, and c be the number of crossing edges. Newman's modularity [38] becomes

$$Q = \frac{V_A(2m - V_A)}{2m^2} - \frac{c}{m} = \frac{1}{2} - \frac{(V_A - m)^2 + 2mc}{2m^2} \tag{G.1}$$

Maximizing Q is equivalent to minimizing the integer energy $E=(V_A-m)^2+2mc$. The MILP formulation uses binary $x_v$, continuous $0 \le y_{uv} \le 1$ for each edge, continuous $-m \le d \le m$, and continuous $0 \le z \le m^2$. It minimizes $z+2m\Sigma_{(u,v)} y_{uv}$ subject to

$$d = \sum_v d_v x_v - m, \quad y_{uv} \ge x_u - x_v, \quad y_{uv} \ge x_v - x_u \tag{G.2}$$

$$z \ge (2k+1)d - k(k+1) \quad (k = -m, \ldots, m-1), \qquad x_0 = 0 \tag{G.3}$$

Although d is continuous in the formulation, binary x makes it integer. At every integer d in [-m,m], the maximum of the consecutive-integer secants equals $d^2$, so minimization sets $z=d^2$. Positive coefficients likewise set $y_{uv}=|x_u-x_v|$. Fixing $x_0=0$ removes only label symmetry. An empty community remains legal and the one-community partition has Q=0.

The reference calculation uses scipy.optimize.milp with bundled HiGHS (SciPy 1.17.1, NumPy 2.4.6, NetworkX 3.6.1), one solver thread and requested relative MIP gap zero. All ten instances terminate with optimal status. For each integer incumbent E*, the numerical dual bound satisfies ceil(dual-$10^{-5}$)=E*, and returned vertex values are integral within $10^{-5}$. Witness Q is recomputed by exact rational arithmetic and checked against NetworkX modularity within $10^{-12}$; the solver objective matches witness energy within $10^{-4}$. The reported validation also checks small path, cycle, complete, disconnected and seeded-random graphs against exhaustive binary assignments, including a one-community optimum. Optimality certification rests on the numerical primal/dual bounds and integer-energy checks; exact rational arithmetic evaluates the witnesses.

Table G.1. Binary-modularity solver-certified references.

| Instance (ca-GrQc-main-) | n | m | Exact rational Q* | E* |
|---|---|---|---|---|
| n50-s0 | 50 | 85 | 83/170 | 170 |
| n50-s1 | 50 | 69 | 1529/3174 | 174 |
| n75-s0 | 75 | 92 | 8159/16928 | 305 |
| n75-s1 | 75 | 139 | 18163/38642 | 1158 |
| n100-s0 | 100 | 226 | 12443/25538 | 1304 |
| n100-s1 | 100 | 180 | 29519/64800 | 2881 |
| n125-s0 | 125 | 231 | 48527/106722 | 4834 |
| n125-s1 | 125 | 199 | 19402/39601 | 797 |
| n150-s0 | 150 | 204 | 10151/20808 | 1012 |
| n150-s1 | 150 | 224 | 48831/100352 | 1345 |

All modularity references use $Q^*=1/2-E^*/(2m^2)$. Astra's high-precision mean gaps are 0.0002611075272071162 for direct construction (8/10 optimal) and 0.012982818923280088 for heuristic-proposal selection (1/10 optimal). GPT-5.4's full-precision objectives yield mean gaps 0.4042398456 and 0.0422861571, respectively, with 0/10 and 1/10 optimal. Optimal-instance counts use a $10^{-10}$ absolute numerical tolerance against the rational reference.

## G.2 Deterministic LT Influence Maximization

Each fixed directed graph induced from SNAP email-Eu-core [37] uses exactly two distinct seeds, activated simultaneously. Every incoming edge of v has weight 1/indegree(v), and each nonseed has threshold 1/2. An inactive nonseed of positive indegree activates exactly when twice its active incoming-neighbor count is at least its indegree. Activation is irreversible; a zero-indegree nonseed never activates spontaneously. The objective f(S) counts all finally active vertices, including both seeds. No random threshold sampling or Monte Carlo estimation is used.

$$f^* = \max_{\{u,v\}\subseteq V,\ u\neq v} f(\{u,v\}), \qquad \text{gap}(\%) = \frac{|f(S)-f^*|}{f^*}\times 100 \tag{G.4}$$

Enumerate every unordered distinct seed pair. An event queue starts with both seeds; each newly active vertex is processed once, incrementing predecessor counts at its outgoing neighbors and enqueuing a neighbor once its threshold is reached. An independent synchronous implementation represents active vertices and incoming-neighbor sets as bitsets, evaluates every threshold against the previous round, and activates eligible vertices together until a fixed point.

Exhaustive enumeration compares the complete final active sets of the queue and synchronous implementations for every pair, not only their sizes. Nonnegative weights and irreversible activation yield the same least closure above the seed set. Exhaustion of all n(n-1)/2 pairs covers the feasible space, so the largest final count is the global optimum; the reported lower and upper bounds both equal it. Queue evaluation costs O(n+m) per pair, giving O(n(n-1)(n+m)/2) overall, plus the independent synchronous validation. The enumeration complexity is specific to the fixed seed count two.

Table G.2. Deterministic LT exact references.

| Instance (email-Eu-core-main-) | n | Edges | Pairs | f* | Witness |
|---|---|---|---|---|---|
| n50-s0 | 50 | 150 | 1225 | 14 | 18,36 |
| n50-s1 | 50 | 128 | 1225 | 17 | 32,34 |
| n75-s0 | 75 | 317 | 2775 | 29 | 59,69 |
| n75-s1 | 75 | 232 | 2775 | 63 | 10,25 |
| n100-s0 | 100 | 348 | 4950 | 25 | 51,65 |
| n100-s1 | 100 | 470 | 4950 | 25 | 53,87 |
| n125-s0 | 125 | 507 | 7750 | 25 | 9,97 |
| n125-s1 | 125 | 643 | 7750 | 16 | 68,114 |
| n150-s0 | 150 | 843 | 11175 | 17 | 29,61 |
| n150-s1 | 150 | 1002 | 11175 | 15 | 20,72 |

Witness seed IDs are zero-based local IDs of the fixed graph, not original SNAP identifiers. A listed pair is one maximizer and need not be unique. These LT optima are the references used for the primary results.